\def\year{2022}\relax
\documentclass[letterpaper]{article} % DO NOT CHANGE THIS
\usepackage{aaai22}  % DO NOT CHANGE THIS
\usepackage{times}  % DO NOT CHANGE THIS
\usepackage{helvet}  % DO NOT CHANGE THIS
\usepackage{courier}  % DO NOT CHANGE THIS
\usepackage[hyphens]{url}  % DO NOT CHANGE THIS
\usepackage{graphicx} % DO NOT CHANGE THIS
\usepackage{natbib}  % DO NOT CHANGE THIS AND DO NOT ADD ANY OPTIONS TO IT
\usepackage{caption} % DO NOT CHANGE THIS AND DO NOT ADD ANY OPTIONS TO IT
\DeclareCaptionStyle{ruled}{labelfont=normalfont,labelsep=colon,strut=off} % DO NOT CHANGE THIS
\usepackage{algorithm}
\usepackage{algorithmic}

\usepackage{newfloat}
\usepackage{listings}
\floatstyle{ruled}
\newfloat{listing}{tb}{lst}{}
\floatname{listing}{Listing}
\usepackage{amsmath, amssymb}
\usepackage{multirow}
\usepackage{array}
\newcolumntype{L}[1]{>{\raggedright\let\newline\\\arraybackslash\hspace{0pt}}m{#1}}
\newcolumntype{C}[1]{>{\centering\let\newline\\\arraybackslash\hspace{0pt}}m{#1}}
\newcolumntype{R}[1]{>{\raggedleft\let\newline\\\arraybackslash\hspace{0pt}}m{#1}}

\title{Globally Consistent Video Depth and Pose Estimation with Efficient Test-Time Training}
\author{
    Yao-Chih Lee,
    Kuan-Wei Tseng,
    Guan-Sheng Chen,
    Chu-Song Chen
}
\affiliations{
    National Taiwan University
}

\usepackage{bibentry}
\begin{document}

\maketitle

\newcommand{\eg}{e.g.}
\newcommand{\etal}{et al. }
\newcommand{\ie}{i.e.}
\makeatletter
\newcommand\notsotiny{\@setfontsize\notsotiny\@vipt\@viipt}
\makeatother
\makeatletter
\newcommand*\subfootnotesize{%
  \@setfontsize\subfootnotesize{7.5}{9.0}%
}
\makeatother

\begin{abstract} % need <= 150 words
%Video dense 
Dense depth and pose estimation is a vital prerequisite for various video applications. 
Traditional %approaches 
solutions suffer from the robustness of sparse feature tracking and insufficient camera baselines in videos. 
Therefore, recent methods utilize learning-based %dense 
optical flow and depth prior to estimate dense depth.
However, %the prior 
previous works require heavy computation time or yield sub-optimal depth results.
%This paper presents 
We present GCVD, a globally consistent method for learning-based video structure from motion (SfM) in this paper. %We introduce %the celebrated 
GCVD %integrates a %keyframe-based 
integrates %an efficient 
a compact pose graph into the 
CNN-based optimization %framework 
to achieve %efficiency and 
globally consistent estimation %with efficiency. %Besides, our 
%based on 
from an effective keyframe selection mechanism. 
It %improves 
can improve the robustness of %the prior 
learning-based methods with flow-guided keyframes and %well-handled 
well-established depth prior.
Experimental results show that %the proposed 
GCVD outperforms the state-of-the-art methods on both depth and pose estimation. %Furthermore, 
Besides, the runtime experiments reveal that %our method performs the fastest in short videos and %still 
it provides strong efficiency in both short- and long-term videos with global consistency provided. 
%We will release our code soon.
   
%\keywords{SfM, video depth and pose estimation, test-time training.}
\end{abstract}

\section{Introduction}

Acquiring %per-frame 
dense depth and camera pose from videos is essential for various applications including augmented reality~\cite{Occlusion2018,Du2020DepthLab}, video frame interpolation~\cite{bao2019depth}, view synthesis~\cite{choi2019extreme,yoon2020novel,liu2021infinite} and stabilization~\cite{liu2009content,lee20213d}. 
In this paper, we propose a learning-based approach to achieve the concurrent inference of scene depth and camera pose from offline videos.
Our study belongs to the research track of SfM with the videos acquired. 
Unlike visual SLAM~\cite{engel2014lsd,mur2015orb,tateno2017cnn,teed2021droid} assuming streaming videos that should be estimated on-line, more information of batched frames stored in the video can be used for a globally consistency estimation. 

%Nevertheless, estimating 
Inferring both depths and %camera 
poses %in a video 
for every frame is essentially a challenging chicken-and-egg problem. 
Traditional solutions rely on the established SfM tools (\eg, COLMAP \cite{schonberger2016structure}) to estimate the camera trajectory and then perform multi-view stereo. %. Besides the limitation of producing
However, the tools often yield incomplete depth and suffer from the robustness issue due to the fragile, sparse feature tracking. 
Especially in the videos containing dynamic objects, motion blur, or large texture-less regions, the estimation process usually fail early even in the pre-processing step.

\begin{figure}
    \centering
    %\includegraphics[width=.5\linewidth]{figures/comparison-summary.png}
    %\begin{tabular}{cc}
        %\includegraphics[width=.42\linewidth]{figures/error_comparison_plot.png} & \includegraphics[width=.48\linewidth]{figures/time_comparison_plot.png} \\
        %(a) Depth and Pose Errors & (b) Runtime Comparison
    %\end{tabular}
    \includegraphics[width=\linewidth]{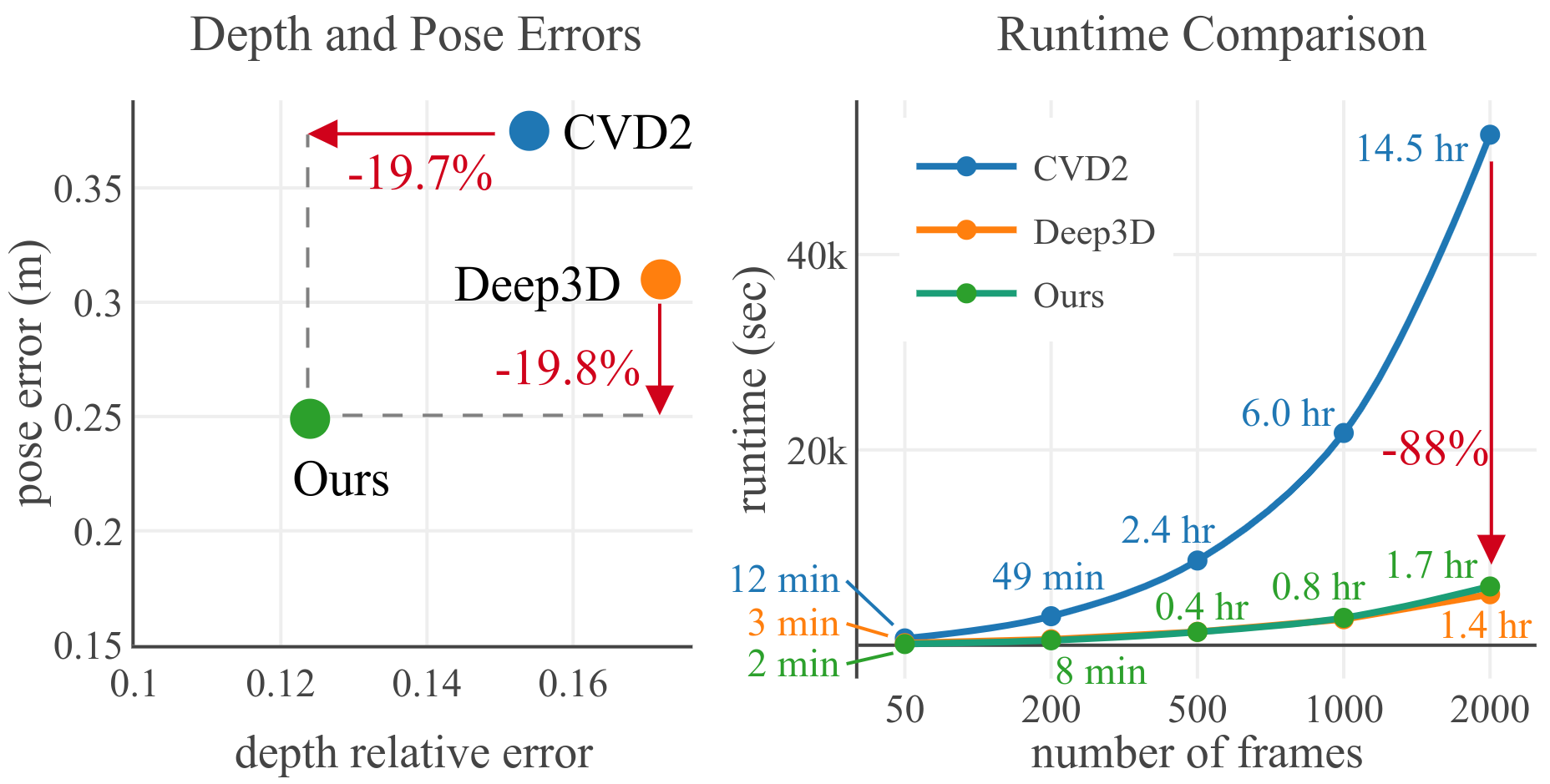}
    \caption{\textbf{Comparison of estimation errors and runtime.} The proposed method outperforms the state-of-the-art methods CVD2~\cite{kopf2021robust} and Deep3D~\cite{lee20213d} on both depth and pose estimation on 7-Scenes dataset~\cite{shotton2013scene}. Furthermore, our method performs the fastest in short videos; and slightly slower than Deep3D in long videos while Deep3D does not maintain global consistency.}
    \label{fig:comparison-summary}
\end{figure}

%With the rapid progress in deep learning, we have witnessed various approaches to estimating dense depth maps and achieving promising results. 

%Nevertheless, most 

In learning-based studies, many works~\cite{eigen2014depth,Ranftl2020,miangoleh2021boosting} focus on estimating the depth using a single image (\ie, single depth estimation). 
Though every single depth is visually plausible in 2D, it is up-to-scale or even a \emph{projective ambiguity}~\cite{hartley2003multiple}.
Thus the individual depths obtained from sequential frames suffer easily from temporal and geometric inconsistency. 
To learn from videos, unsupervised (or self-supervised) methods~\cite{zhou2017unsupervised,yin2018geonet,godard2019digging,bian2019unsupervised} are proposed. They may maintain the scale consistency~\cite{bian2019unsupervised} but %often collapse when training on wild videos due to the ill-posed constraints. 
%Besides,
their capability of generalizing the pre-trained model to the testing scene is usually %highly 
restricted by the training data.

To overcome these issues, \textit{test-time training} video-based SfM methods are introduced~\cite{luo2020consistent,kopf2021robust,lee20213d}, which optimize the depth and pose of a test video directly. %and often ensure \emph{temporary consistency}.
The solutions provide a promising way to maintain the depth and ego-motion coherence in an input video.
Luo~\etal propose CVD~\cite{luo2020consistent} to refine the single-depth network for enhancing depth consistency of a monocular video. However, it relies on the COLMAP tool in advance to obtain the camera poses and thus is still limited by the robustness issue of traditional SfM. 
Kopf~\etal extend CVD to CVD2~\cite{kopf2021robust} without requiring SfM tools. 
CVD2 performs simultaneous depth and pose optimization with the off-the-shelf single depth network, MiDaS~\cite{Ranftl2020}, as initial. 
An issue is that the optimization is often affected by the bias of single depths and therefore yields poor pose results. 
Lee~\etal present Deep3D~\cite{lee20213d} to learn depth and pose of the test video to solve video stabilization problem. 
However, the optimization still results in sub-optimal estimations for short-camera-baseline videos due to the lack of single depth prior. 
Moreover, CVD2 and Deep3D ensure temporal consistency in a local range, whereas neither of them consider global consistency which is crucial especially in long videos.

\begin{figure*}[t]
    \begin{center}
	\includegraphics[width=\linewidth]{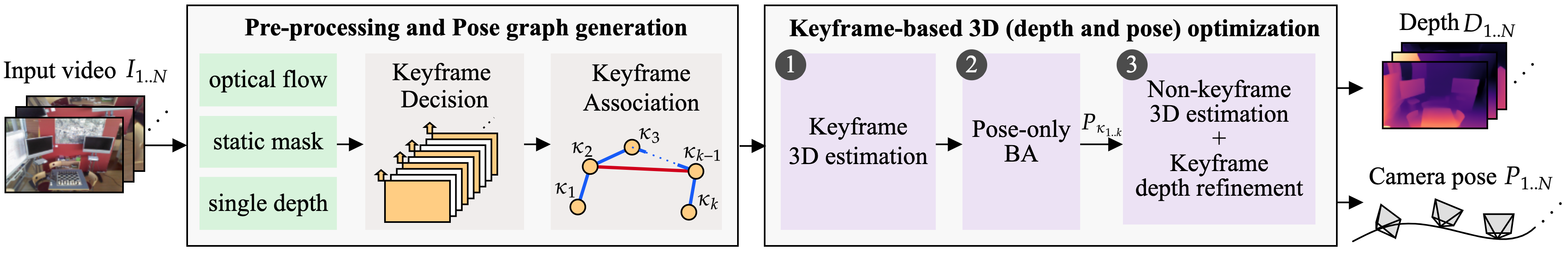}
	\end{center}
	\caption{\textbf{Pipeline overview.} GCVD takes a monocular RGB video as input to estimate per-frame depth and pose. The pre-processing and pose graph generation stage (Sec.~\ref{sec:pre-process}) first estimates adjacent optical flows and find static masks to filter likely-dynamic regions. It then uses single depths as prior for the optimization stage and constructs a pose graph by selecting the keyframes as vertices. The %connecting 
	sequential (blue) and non-sequential (red) edges are thus formed via keyframe association. The optimization stage (Sec.~\ref{sec:keyframe-based-optimization}) takes the estimations in the pre-processing stage as priors to optimize the depth and pose of input frames using the same Joint Optimization of Depth and Pose (Sec.~\ref{sec:joint-optimization}). The keyframe %optimization 
	3D estimation is followed by pose-only bundle adjustment (BA) for global consistency. The depth of keyframes and depth/pose of the remaining non-keyframes can then be optimized more efficiently through learning.
	}
	\label{fig:pipeline}
\end{figure*}

\noindent\textbf{Global-consistency of video depth}: We argue that current test-time training methods lack of the ability to achieve global consistency between the depth and pose, which is a demanding issue for accurate and long-term SfM estimation and thus they will yield drifting outcomes for the input videos. 
%This paper proposes 
We propose GCVD, a globally consistent test-time training method for depth and pose estimation based on offline videos. 
GCVD has the advantage that it can achieve robust inference without relying on traditional SfM tools.
Our method %utilizes 
integrates a keyframe-based pose graph into learning to attain the global consistency. % efficiently.
Fig.~\ref{fig:pipeline} illustrates its pipeline.
In our method, the keyframes are extracted from the input video to compose a pose graph with sequential %edges 
and associated non-sequential edges.
It estimates the depth and pose of keyframes and performs %\textcolor{blue}{\sout{loop closure with }}
pose graph optimization %(PGO) 
to fulfill the global consistency. 
%Accordingly, 
Then, the depth and pose of the remaining frames %(\ie, non-keyframes) 
can be estimated efficiently leveraging the keyframes.
The experimental results validate the strength of GCVD (Fig.~\ref{fig:comparison-summary}) that our approach outperforms the state-of-the-art approaches, CVD2 and Deep3D, on both depth and pose estimations (7-Scenes dataset~\cite{shotton2013scene}). 
Besides, it is significantly faster than  CVD2 and achieves a good trade-off between efficiency and global consistency in contrast to Deep3D. 
%In sum, the 
The main characteristics of our GCVD include:

\noindent\textbf{Global consistency}: %\textcolor{purple}{To our best knowledge, we %integrate keyframe-based pose graph with loop closure into our joint depth and pose %optimization
%learning framework to attain global consistency with robustness.
%introduce the first joint depth and pose learning framework with keyframe-based pose graph and loop closure to attain global consistency with robustness.}
To tackle the challenging on joint depth and pose estimation based on video collections, to our knowledge, we introduce the first %learning 
test-time-training method %framework %with \textcolor{blue}{keyframe-based pose graph optimization} 
that enforces the global consistency with robustness.

%\noindent\textbf{Flow-guided keyframe strategy for robustness and efficiency}:
%We leverage the adjacent optical flow to adaptively extract the keyframes with proper camera baseline for robust 3D optimization as initials. Subsequently, the non-keyframes can be efficiently estimated with the keyframes' estimation.

\noindent\textbf{%Proper baselines and 
Efficient global-pose-graph and optimizer}:
Our method %can choose automatically the keyframes that guarantee proper baselines for robust SfM. It 
requires merely the pose-only bundle adjustment on the keyframes, and can then leverage network learning for estimating the depths and poses for all frames efficiently.

\noindent\textbf{Performance improvements and competitive speed}: % high efficiency}: 
Our method outperforms SOTA on both depth and pose with over 19\% improvement on 7-Scenes dataset~\cite{shotton2013scene}, and also shows strong computational efficiency. %with the %keyframe-based 
%CNN optimization 
%learning framework.

\section{Related Work}
\noindent\textbf{Traditional Approach:}
Traditional %Structure-from-Motion (SfM) 
SfM \cite{wu2011visualsfm,moulon2016openmvg,schonberger2016structure} jointly estimate the 3D %scene 
structure and camera poses of multi-view images via bundle adjustment \cite{ceres-solver,kummerle2011g} on the local features. 
Subsequently, multi-view stereo (MVS)~\cite{furukawa2015mvstutorial} %algorithms follow 
following the estimated pose obtain dense depth but often yield holes and noises in general. 

In another track, traditional visual odometry (VO) and SLAM~\cite{klein2007parallel,engel2014lsd,mur2015orb} usually maintain keyframes and a pose graph to perform efficient and consistent localization. 
Besides, %\sout{loop detection and closure~\cite{mur2014fast}} %are often 
bundle adjustment (BA)~\cite{triggs1999bundle} and pose graph optimization~\cite{kummerle2011g}
can be introduced to prevent drifts and enhance global consistency. 

%Nevertheless, the 
The %pose and depth 
performance of traditional methods generally rely on successful feature tracking. 
The sparse local features~\cite{lowe2004distinctive,bay2006surf,rublee2011orb} extracted are often fragile in various challenging conditions such as %large %texture-less 
homogeneous areas, motion blur, and illumination changes. 
Hence, they are %usually 
demanding to obtain dense depth maps and non-robust enough to handle the texture-less and blur situations for the videos in the wild.

\noindent\textbf{%Depth or Multi-view Supervision with Known Pose:
Learning Depth Only (Given Pose):}
%Supervised learning has been widely used for the ill-posed single depth estimation problem~\cite{eigen2014depth,liu2015learning,eigen2015predicting}. %Nevertheless, 
%A key demand %challenge 
%is to acquire sufficient real-world groundtruth depths with LiDAR~\cite{saxena2008make3d,Geiger2013IJRR} or RGB-D cameras~\cite{sturm12iros,Silberman:ECCV12,shotton2013scene}. %dai2017bundlefusion} 
%that are either expensive or can obtain incomplete data only. 
%To tackle the issue, the %approaches 
Supervised learning has been widely used for the ill-posed single depth estimation problem~\cite{eigen2014depth,liu2015learning,eigen2015predicting}. Apart from acquiring real depth as groundtruth, some methods \cite{garg2016unsupervised,godard2017unsupervised,gonzalez2021plade} learn the single depth with binocular %image 
pairs. % via self-supervision. % in a self-supervised manner. %Another 
Other studies % approaches %rely on
leverage synthetic %datasets~
data~\cite{mayer2016large} or pseudo groundrtuth %depth~
\cite{chen2016single,li2018megadepth,chen2020oasis,li2019learning}.
MiDaS~\cite{Ranftl2020} %further 
obtains relative depth of stereo pairs from large-scale and diverse 3D movies. Recently, Miangoleh~\etal\cite{miangoleh2021boosting} integrate multi-scale depth of MiDaS %depth 
to handle high-resolution images. 
Yin~\etal\cite{yin2021learning} utilize point cloud networks to %tackle 
solve the perspective ambiguity of a single depth. 
Despite the single-depth methods show visual plausibility on individual depth maps, the issue of geometrical inconsistency among multi-views is not addressed. 

With known camera poses, learning-based multi-view stereo can estimate dense depth of multiple images. % with known camera poses. 
In \cite{huang2018deepmvs,yao2018mvsnet,im2019dpsnet}, plane sweep algorithm is employed to estimate dense depth in a supervised manner. 
The methods in \cite{long2021multi,wimbauer2021monorec} estimate video depth with known camera pose or obtain the pose via COLMAP. 
%Nevertheless, COLMAP is extremely time-consuming and unsuitable for long videos as the time complexity tends to grow exponentially with views due to bundle adjustment.
%In contrast, our goal is to jointly estimate consistent dense depth and camera poses of a video without COLMAP due to the robustness and time-consuming issues.

%\noindent\textbf{Learning-based Pose Estimation:}
%Many studies adopt supervised learning to %visual-based 
%camera pose estimation. The camera re-localization methods~\cite{kendall2015posenet,brachmann2017dsac} learn to infer the pose of a single image in a known 3D model. %In contrast, other 
%Other supervised approaches have been proposed to learn multi-view relative pose~\cite{zhuang2021fusing} and visual odometry~\cite{wang2017deepvo,wang2018end,xue2019local,xue2019beyond}. 
%However, acquiring the groundtruth pose is challenging in the real world. In contrast, our method estimates the camera trajectory and depth of a video without groundtruth pose.

%\subsection{Joint Learning in Depth and Pose Estimation}
\noindent\textbf{%Supervised Learning of Depth and Pose: 
Depth and/or Pose Supervision and Joint Estimation%with Unknown Pose in Testing
:}
Approaches of this type utilize groundtruth depths and/or known-pose views for training; the obtained joint-depth/pose estimator is then applied to the test-scene sequences of unknown poses.
%Some supervised methods~\cite{ummenhofer2017demon,zhou2018deeptam,wei2020deepsfm,wang2021deep} utilize solid SfM prior to explore joint learning of multi-view depth and pose estimation. 
For depth estimation, UNet-like models are used to perform per-pixel depth regression~\cite{ummenhofer2017demon,tateno2017cnn,zhou2018deeptam,bloesch2018codeslam,czarnowski2020deepfactors}.
Deep cost volume with plane sweep structure are also adopted~\cite{wei2020deepsfm,teed2018deepv2d,wang2021deep}.
To estimate the pose, some methods directly use networks for regression~\cite{ummenhofer2017demon,zhou2018deeptam,wei2020deepsfm,teed2021droid};
some others leverage the fundamental multi-view and epipolar geometry~\cite{tateno2017cnn,bloesch2018codeslam,teed2018deepv2d,czarnowski2020deepfactors,wang2021deep}.
%Recent methods~\cite{wei2020deepsfm,wang2021deep,teed2018deepv2d} adopt some 3D representation structure (\eg, plane sweep volume) to the depth estimation branch.
%The pose branches exploit regression networks~\cite{ummenhofer2017demon,zhou2018deeptam,wei2020deepsfm} or fundamental epipolar geometry~\cite{wang2021deep} to estimate the camera pose. 
%Supervised SLAM approaches~\cite{bloesch2018codeslam,teed2018deepv2d,czarnowski2020deepfactors,teed2021droid} \textcolor{blue}{also require depth or pose supervision during training.
%(combined to the previous. It seems that the title of [71] is DeepBundleAdjustment, which is highly related to our work?)} %have introduced pose graph optimization and full bundle adjustment into the systems to enhance global consistency. %Nonetheless, to our knowledge, the test-time training-based video SfM approach~\cite{kopf2021robust,lee20213d} has not taken global consistency into account.
However, acquiring the groundtruth pose or depth in real world is non-trivial %challenging and expensive 
(\eg, requiring LiDAR~\cite{saxena2008make3d,Geiger2013IJRR} or RGB-D cameras~\cite{sturm12iros,Silberman:ECCV12,shotton2013scene}).
The demanding ability of generalization to the testing data of unknown scenes also restricts the usage in practice.

\noindent\textbf{Self-Supervision for Depth and Pose:}
Methods in this category learn a joint depth/pose inference network in a self-supervised (or unsupervised) manner without relying on pre-given groundtruths~\cite{zhou2017unsupervised,gordon2019depth,godard2019digging,chen2019self,li2019sequential,bian2019unsupervised,ranjan2019competitive,zou2020learning,jiao2021effiscene,watson2021temporal,Zhou_2021_ICCV,Jung_2021_ICCV}. 
They cooperatively regulate the depths and %camera ego-motion 
poses on monocular videos via warping-based view synthesis from the training data. 
Further approaches use optical flow~\cite{yin2018geonet,ranjan2019competitive,jiao2021effiscene} and segmentation~\cite{casser2019unsupervised,ranjan2019competitive} to improve the performance and handle dynamic objects.
Yet, in test time, most %depth
networks still perform single-depth estimation and thus yield incoherent results.
%Besides, these studies %still 
The methods also suffer %easily 
from the generalization issue due to the domain gap between training and testing data.

\noindent\textbf{Test-Time Training Approach:} 
To avoid the difficulty of applying the inference model learned from given environment to unseen environment, test-time-training methods have been proposed more recently, making the solutions better for practical use.
Despite the speed of test-time-training is slower than pure inference, the approaches meet the requirement of offline videos that need no real-time processing.
CVD~\cite{luo2020consistent} is the first attempt toward 
test-time %optimization 
training to obtain geometrically consistent depth and pose estimation of a video.
However, CVD relies on the SfM tool COLMAP which suffers from the fragile and sparse reconstruction and the computation is slow.
Deep3D~\cite{lee20213d} uses a network-based optimization framework to learn video depth and pose for video stabilization, but the depth performance is restricted % by the depth prior adopted.
due to the lack of single depth prior.
CVD2~\cite{kopf2021robust} uses a deformative mesh on single depth with a post-processing filter to show promising depth results; however, the pose performance still suffers from the bias of single depth as no fine-tuning mechanisms have been accommodated for the depth net.
Besides, CVD2 requires tremendous optimization time.
Its quantitative performance is only evaluated on the videos of 50 frames~\cite{Wulff:ECCVws:2012}. % with the traditional optimizer~\cite{ceres-solver}.

%Deep3D and CVD2 
Current works only ensure local consistency with nearby frames, whereas the global consistency is not tackled. 
We introduce GCVD that is the first test-time-training method with global consistency, which can obtain more accurate and robust results. 
%Our work integrates efficient keyframe-based pose graph %\sout{loop closure}
%optimization and a CNN-based optimization framework to 
Due to the global consistency, GCVD is scalable to long videos. %It can handle long videos (%\eg, such as 1000 frames). %, and the computational efficiency is even higher.
In our experiments, 1000-frames videos are used to validate the performance.

\section{Methodology}
Given an $N$-frame video $I_{1..N}$, our goal is to estimate dense depth maps $D_{1..N}$ and camera poses $P_{1..N}$. %  without any traditional SfM tool %(\eg, COLMAP~\cite{schonberger2016structure}) 
%in advance due to the issues of robustness and time complexity. The 
Our CNN-based optimization framework utilizes the %solid structure-from-motion (SfM) 
learning-based SfM to jointly estimate depth and pose of each frame. % with multiple frames. 
Nevertheless, applying SfM to a video may suffer from various challenges such as small camera baseline motion and lacking of co-visible scenes among frames. 

\noindent\textbf{Proper baselines arrangement}: %Prior state-of-the-art 
Current test-time-training solutions (\eg, Deep3D~\cite{lee20213d} and CVD2~\cite{kopf2021robust}) simply take near-to-distant %neighboring 
neighbor frames with certain frame intervals empirically selected (\eg, 1, 2, 4, 8)  to ensure a coherence %and sufficient 
baseline. 
%Yet, they empirically select the frames  which 
The strategy does not take the real disparities between the frames into account, and does not guarantee the proper baselines among the frames for SfM. %Consequently, 
In our work, we leverage the recent progress of deep optical flow estimation, and introduce dense flow-guided keyframes to perform initial optimization with adequate baseline motions. 
The pipeline (Fig.~\ref{fig:pipeline}) is introduced in \textit{Pre-processing and Pose Graph Generation} %Sec.~\ref{sec:pre-process} 
and \textit{Keyframe-based 3D Optimization} %\ref{sec:keyframe-based-optimization} 
and the network-based optimization module is elaborated in \textit{Joint Optimization of Depth and Pose}. %Sec.~\ref{sec:joint-optimization}.

%\subsection{Pipeline}\label{sec:pipeline}
\subsection{Pre-processing and Pose Graph Generation}\label{sec:pre-process}

%\subsubsection{Pre-processing}\label{sec:preprocessing}
%To overcome the robustness issue of sparse feature in traditional approaches, 
Videos often contain motion blurs and large view-direction changes, yielding failures of traditional sparse features
on matching. %Thus, 
We use learning-based optical flows %estimation 
among video frames %is used 
to overcome this difficulty and obtain dense flow maps.
%in the optimization framework. 
However, in contrast to existing approaches CVD2 and Deep3D which need the computation of flows from a target frame to many %multiple %nearby 
other frames, our %approach 
GCVD initially estimates the depth and pose of keyframes with reliable camera baselines and then enforces global optimizations later; thus, we can only take the flow of adjacent views (\ie, $\Hat{F}^t_{t\pm1}$) to save the computation burden.

Besides, to prevent dynamic objects from disturbing pose estimation, semantic segmentation (\eg, Mask-RCNN in Detectron2~\cite{wu2019detectron2}) is used to obtain a binary mask $\Hat{M}_t$ that filters out the likely-dynamic pixels in each frame $I_t$. %In addition, we 
We also take the single depths of MiDaS~\cite{Ranftl2020} as the depth prior to regularize the %joint depth and pose 
optimization.

%\noindent \textbf{Static mask.} The dynamic objects in the scene
%Similar to CVD~\cite{luo2020consistent} and CVD2~\cite{kopf2021robust}, the binary masks $\Hat{M}_{1..N}$ filter out the likely-dynamic pixels according to the categories inferred by Detectron2~\cite{wu2019detectron2}.

%\noindent \textbf{Single depth estimation.} Although the single depth network~\cite{Ranftl2020} suffers from scale inconsistency among image sequences, we leverage the single depths $\Hat{D}_{1..N}$ as the depth prior to regularize the joint depth and pose optimization.

%\subsubsection{Pose graph generation leveraging learning}\label{sec:pose-graph}
%Though the 
The ideas of keyframe and pose graph have been widely used in traditional SLAM~\cite{klein2007parallel,engel2014lsd,mur2015orb} and SfM~\cite{schonberger2016structure,barath2021efficient} to reduce the computational complexity and ensure global consistency for large-scale reconstruction. %, the current SOTA solutions on test-time-training video SfMs~\cite{lee20213d,kopf2021robust}  do not maintain a global structure of universal views and thus fail to maintain the global consistency yet.
Our method constructs a pose graph. %,while using
Nevertheless, unlike traditional SLAM or SfM, we use learning-based approach
to provide better robustness in the optimization. % on motion blurs and view changes. %for initial estimation to achieve efficient and global consistency for long sequences. 
The pose graph %consists of 
has $k$ keyframes ($\kappa_{1..k}$) as its vertices, % extracted by an adaptive keyframe decision strategy instead of uniform selection.
and the edges of the graph include the 
%sequential edges and associated non-sequential %associated 
%edges.
sequential edges and the non-sequential co-visible edges, as depicted below.
 %The sequential edges connect multiple temporally-nearby keyframes to ensure local consistency. For global consistency, each non-sequential edge connects a pair of temporally-distant keyframes that share a common scene via similarity association.

\noindent \textbf{Dense-flow-guided keyframe decision.}
To sample representative keyframes from videos, traditional solutions rely on sparse feature tracking while long feature tracks are challenging to obtain. 
Instead, the dense optical flow acquired in the pre-processing can provide a reliable reference for keyframe decision.
Thus, the frame $I_t$ is chosen as a keyframe if the accumulated flow magnitude $\Bar{F}^{\kappa^{*}}_{t}$ from the last selected keyframe $\kappa^{*}$ exceeds a %displacement 
movement threshold $\Delta$.
\begin{equation}
    \Bar{F}^{\kappa^{*}}_{t} = \sum^{t-1}_{i=\kappa^{*}}\left( \frac{1}{|\Hat{M}_i|}\sum_{x\in \Hat{M}_i}\|\Hat{F}^i_{i+1}\|_2 \right),
\end{equation}
where we set $\Delta=0.1$ and only %consider the displacement 
use the static regions $\Hat{M}$ for evaluation. %Subsequently, the adjacent 
Then, the flow of adjacent keyframes, $\Hat{F}^{\kappa_i}_{\kappa_{i\pm 1}}$, %among sequential keyframes 
are established %computed %as in Sec.~\ref{sec:preprocessing} 
for the keyframe optimization. 
As we use the accumulated adjacent flows to pick just-needed frames, compared to uniform selection, a more compact and exemplary keyframe set can be built.
After selecting $k$ keyframes, the sequential edges are formed by connecting the keyframes of nearby indices within a subset of $\kappa_{i \pm \alpha}$
%the sequential and covislble non-sequential edges connect the keyframes (\ie, vertices) 
in the pose graph %The sequential edges link multiple temporally-nearby keyframes to
to ensure local consistency.
For those keyframe pairs with the index differences exceeding $\alpha$, we will consider their co-visibility of shared scenes to form the further edges of non-sequential co-visible views to enforce better the global consistency.
%For global consistency, each non-sequential edge connects a pair of temporally-distant but 3D-spatially-nearby keyframes that share a co-visible scene. %via association.

\noindent \textbf{CNN-based keyframe association.} 
%To associate non-sequential edges between %keyframes that each associated pair share a common scene,
%To link the temporally-distant keyframe pair sharing a co-visible scene, the 
The keyframe pairs with high image similarities are picked %for further geometric verification 
and geometrically verified to serve as non-sequential co-visible edges.
%traditional approaches exploit bag-of-words \cite{sivic2003video,galvez2012bags} yet are memory-inefficient to maintain visual vocabularies. 
%By contrast, the CNN-based feature extraction is able to match frames~\cite{zhang2017loop,arshad2021role}. %We measure the cosine similarity of the image-level features extracted by a ImageNet-pretrained ResNet~\cite{he2016deep} encoder with global average pooling and $L2$-normalization. Hence, a similarity matrix $A_{k\times k}$ is formed and then the loop candidates are sampled from $A$ by a similarity threshold $\delta$ ($=0.9$) and max-pooling for non-maximum suppression.
%In our approach
We leverage the deep features of the $k$ keyframes and compute their cosine similarity one-by-one to form a similarity matrix $A_{k\times k}$. 
%To measure the similarity $A_{a,b}$ (=$A_{b,a}$) between $\kappa_a$ and $\kappa_b$, we first extract the features of the two keyframes using an ImageNet-pretrained ResNet encoder~\cite{godard2019digging}. 
%The output features are passed through a global average pooling layer along with $L2$-normalization. %Thus, the similarity $A_{a,b}$ is the inner-product of the two normalized image-level features. 
For each keyframe, the feature is extracted by an ImageNet-pretrained ResNet encoder~\cite{godard2019digging}. The output feature is then passed through a global average pooling layer and $L2$ normalization. Thus, $A_{k\times k}$ can be simply computed by the inner-product of the $k$ normalized feature vectors.
Then, the associated pairs are sampled from $A$ by a similarity threshold $\delta$ ($=0.9$) and max-pooling for reducing redundant pairs.

%Besides, the geometric verifications are necessary to filter the noisy loop candidates and ensure that %the 
%every keyframe pair
%shares a common scene. Traditional verifications utilize the inlier ratio of estimated fundamental matrix or homography via %sparse features~\cite{lowe2004distinctive,bay2006surf,rublee2011orb} 
%SIFT~\cite{lowe2004distinctive} and RANSAC~\cite{fischler1981random}, while it is not reliable enough. We additionally examine the loop candidates with the forward-backward consistency of the dense optical flow.
%Moreover, in order to guarantee adequate co-visible areas for the optimization %of   chusong:
%by loop pairs, the candidates are removed if the average flow magnitude exceeds the %displacement
%movement threshold $\Delta$ of keyframe decision.
Besides, the geometric verifications are necessary to filter the noisy associated keyframe pairs. Traditional verifications utilize the inlier ratio of estimated fundamental matrix or homography via SIFT~\cite{lowe2004distinctive} and RANSAC~\cite{fischler1981random}, while it is not reliable enough. We additionally examine the forward-backward consistency of the dense optical flow of each associated pair. Moreover, to guarantee adequate co-visible areas for optimization, the associated pairs are removed if the average flow magnitude exceeds the movement threshold $\Delta$ of keyframe decision.

%the loop pairs are selected if the further geometric verifications are all satisfied:
%\begin{itemize}
%    \setlength\itemsep{0em}
%    \item The inlier ratio of fundamental matrix or homography via SIFT~\cite{lowe2004distinctive} and RANSAC~\cite{fischler1981random} is greater than 0.5.
%    \item The ratio of forward-backward consistency of the estimated dense optical flow is greater than 0.1.
%    \item The average flow magnitude is lower than the displacement threshold $\Delta$ of keyframe decision.
%\end{itemize}

\begin{figure}[t]
    \begin{center}
	\includegraphics[width=\linewidth]{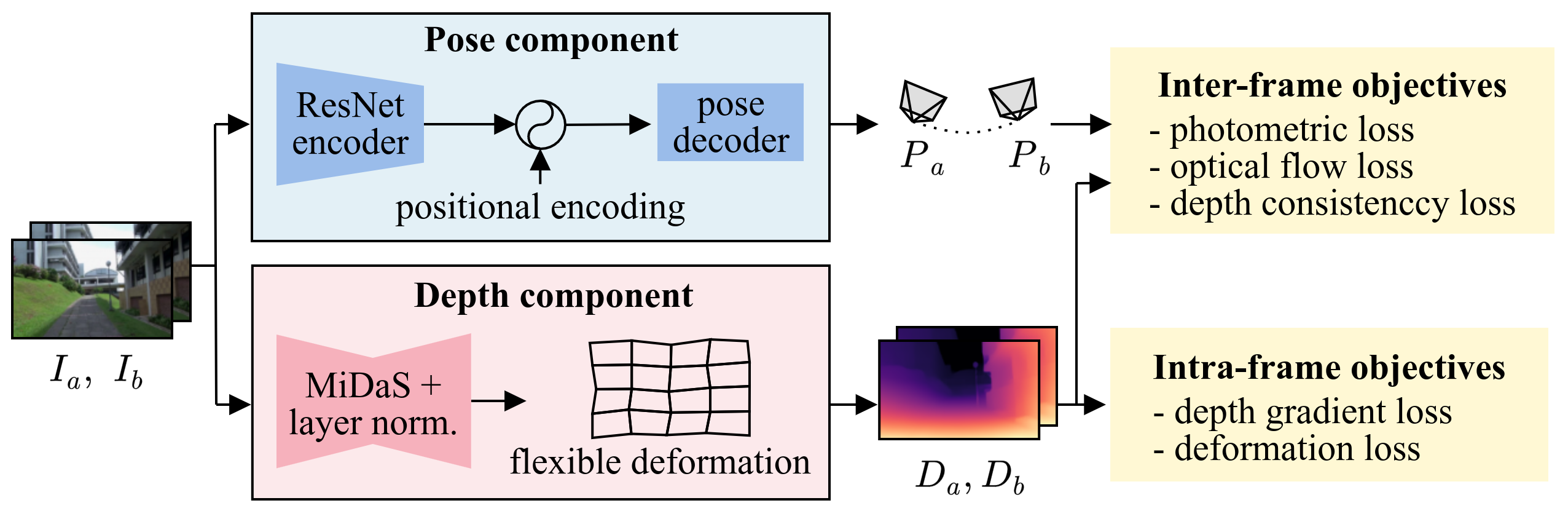}
	\end{center}
	\caption{\textbf{Joint optimization of depth and pose.}
	%The optimization takes a set of frames as input to estimate the per-frame depths and poses. In \emph{Keyframe optimization}, $\kappa_{1..k}$ are taken as inputs. Then $I_{1..N}$ are passed with frozen $P_{\kappa_{1..k}}$ in \emph{Non-keyframe optimization}.}
	%Both depth and pose modules take a single RGB frame as input to output estimations with multiple objectives.
	The 3D optimization module takes a set of frame pairs as input to estimate the depths and poses.
	The depth component adopts MiDaS~\cite{Ranftl2020} with an additional layer normalization and a learnable mesh deformation~\cite{kopf2021robust}. 
	The pose component consists of ResNet-based encoder-decoder~\cite{godard2019digging} with a positional encoding~\cite{vaswani2017attention} to encourage better convergence.
	Both Steps 1 and 3 in the Keyframe-based 3D Optimization stage of Fig.~\ref{fig:pipeline} use this module for learning. %each edge in the pose graph serves as a frame pair for the optimization. Then, $I_{1..N}$ are passed with frozen global pose $P_{\kappa_{1..k}}$ in  (Fig.~2)
	}
	\vspace{0cm}\vspace{-0.4cm}
	\label{fig:modules}
\end{figure}

\subsection{Keyframe-based 3D Optimization}\label{sec:keyframe-based-optimization}
%After obtaining the pose graph, we perform the optimization of the keyframes sequentially.
The proposed pipeline performs keyframe optimization with suitable camera movement to achieve robust estimation as initials. Then, pose graph optimization is utilized to retain global consistency of keyframes' pose estimations. Finally, the depth and pose of the remaining frames are estimated efficiently according to the %computed 
pose graph obtained.
In the following, we give an overview of the procedures. Details of our %3D 
joint depth-and-pose deep network model %optimization %framework
is depicted in \textit{Joint Optimization of Depth and Pose}. %Sec.~\ref{sec:joint-optimization}.

\noindent \textbf{Keyframe %optimization
3D estimation via deep model.} 
For each keyframe $\kappa_i$, the depth and pose are learned with multiple nearby keyframes of $\kappa_{i\pm\tau}$ (\ie, the sequential edges in pose graph) with a descending weight $\frac{1}{\tau}$ to acquire consistent results. Besides, a mini-batch of keyframes are optimized simultaneously %at a time 
with a GPU. Thus, %the fragments of 
%all keyframes in a mini-batch 
the mini-batches are overlapped with the %maximum 
interval $\tau$ to ensure coherent solutions~\cite{lee20213d}.
In our work, we set $\tau \in \{1,2,4,8\}$. % in the implementation.
The relative poses obtained for sequential edges are then recorded for the next pose-only bundle adjustment.
%Additionally, 
Likewise, we optimize the non-sequential co-visible keyframe pairs (without decreasing weights) and obtain the their relative poses accordingly. Hence, each edge of the pose graph is set up with an initial relative-pose transformation.

%\noindent \textbf{Loop closure with efficiency.}
%Our GCVD first optimizes the detected non-sequential loop pairs. Hence, each edge of the pose graph has been %constructed 
%established with an initial relative pose transformation. %Subsequently, 
\noindent\textbf{Global pose graph optimization with efficiency.}
The pose graph optimization~\cite{kummerle2011g} is used to refine the pose estimations of keyframes from the above initialization. Note that we perform \emph{pose-only} bundle adjustment for $P_{\kappa_{1..k}}$ rather than depth and pose bundle adjustment for efficiency. Instead, the depths of keyframes are fine-tuned %with the optimization of non-keyframes 
in the next step leveraging the %deep 
learning models.
The extensive bundle-adjustment overhead can thus be shared with deep networks that are cooperated %CNNs %that are %co-used
to yield a more %accurate and   chusong:
efficient optimization.

\noindent \textbf{Non-keyframe 3D optimization and keyframe depth refinement.%via deep models.
} 
%Lastly, 
Besides fine-tuning the keyframe depth, the remaining non-keyframes are optimized with the fixed keyframe poses (\ie, $P_{\kappa_{1..k}}$) over fewer iterations %using 
via the deep network. Likewise, $(D_t, P_t)$ of each frame is optimized with multiple nearby views $t\pm\tau$. %In the end, 
We then obtain the depth $D_{1..N}$ and pose $P_{1..N}$ of the entire video. % $I_{1..N}$. %are then accomplished.

\subsection{Joint Optimization of Depth and Pose}\label{sec:joint-optimization}
In this section, we introduce the optimization %framework 
module for 3D estimation. % in detail. 
%In contrast to %CVD2  chusong:?? too many CVD2 in contrast 
%using a traditional optimizer, %we take advantage of the neural 
Our approach leverages the deep networks to save the scene information in the %CNN 
model weights %for boosting 
to facilitate sequential fragments optimization.
The module takes a set of frame pairs as input and estimate the depths and poses simultaneously.
For simplicity, we depict the module with a pair of images $(I_a, I_b)$ as input.
More pairs simply use the sum of respective loses.
The networks learn both depth and pose %(Sec.~\ref{sec:modules}) 
to obtain the the output 
$D_a$, $D_b$, $P_a$, $P_b$ with the objectives designed below.
% (Sec.~\ref{sec:objectives}).% for inter-frame consistency (Sec.~\ref{inter-frame-loss}) and intra-frame regularization (Sec.~\ref{intra-frame-loss}).

%\subsubsection{Depth and pose modules}\label{sec:modules}
\noindent\textbf{Depth and pose components.}
As shown in Fig.~\ref{fig:modules}, the depth and pose components estimate the individual depth $D_t$ and 6-DoF global pose $P_t$, respectively, associated with an input RGB frame $I_t$. 
The depth component exploits a MiDaS-pretrained network~\cite{Ranftl2020} with an additional layer normalization to stabilize the output scale of depth estimation. Then, a learnable mesh deformation~\cite{kopf2021robust} is adopted to achieve better alignments among sequential depths.
%followed by a learnable mesh deformation~\cite{kopf2021robust} to achieve better alignments among sequential depths.
%\textcolor{blue}{Besides, we found that inserting an additional layer normalization into MiDas network can stabilize the output scale of depth and facilitate better convergence.}
Unlike CVD2~\cite{kopf2021robust} that directly takes fixed single depths as initials, we %find that 
use a trainable depth network that can refine the bias of the initial depth to encourage better estimations. 
While due to the time and space efficiency, only the last two convolutional layers of MiDaS network are %learnable 
used to learn a larger span of frames at a time.

\begin{figure}[t]
    \begin{center}
	\includegraphics[width=\linewidth]{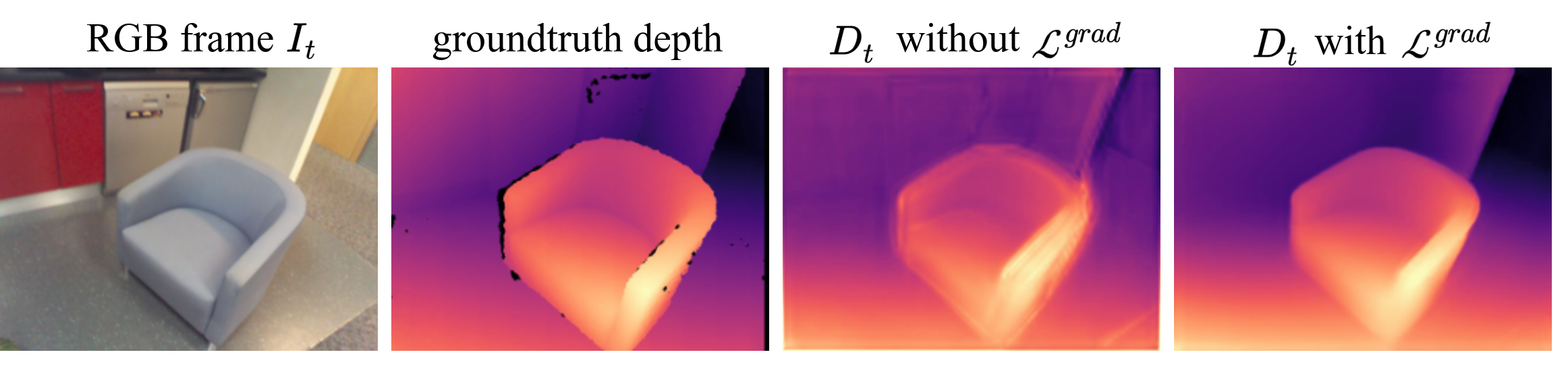}
	\end{center}
	\caption{\textbf{Depth gradient loss $\mathcal{L}^{grad}$} retains the depth prior of initial single depth~\cite{Ranftl2020}. The result without $\mathcal{L}^{grad}$ may introduce blurring after updating the depth component.}
	\vspace{0cm}\vspace{-0.4cm}
	\label{fig:loss_depth_grad}
\end{figure}

The pose component adopts the PoseNet~\cite{godard2019digging} based on ResNet encoder~\cite{he2016deep}. Similar to the depth component, the pose encoder is frozen with ImageNet-pretrained weights and only the decoder can be optimized. The information saved in the learned weights of the decoder can help boost the convergence for the next optimization. Moreover, the feature map is added with a positional encoding~\cite{vaswani2017attention} which encodes the chronological order of the entire sequence %so as 
to enhance the learning %of 
for the sequence. 
%Hence, the optimization can converge faster and even lower error in non-keyframe optimization (Fig.~\ref{fig:loss_pe}).

\begin{table*}[t]
    \centering
    \small
    \begin{tabular}{lcR{1.2cm}R{1.1cm}R{1.1cm}R{1cm}R{1cm}R{1.4cm}R{1.3cm}}
    \hline
    \multicolumn{1}{c}{\multirow{3}{*}{Method}} & \multirow{3}{*}{\begin{tabular}[c]{@{}c@{}}known\\ pose?\end{tabular}} & \multicolumn{4}{c}{Depth Metrics} & \multicolumn{3}{c}{Pose Metrics} \\
    \multicolumn{1}{c}{} &  & \multirow{2}{*}{{AbsRel$\downarrow$}} & \multirow{2}{*}{{SqRel$\downarrow$}} & \multirow{2}{*}{{RMSE$\downarrow$}} & {$\delta<$} & {ATE} & \small{RPE Trans} & \small{RPE Rot} \\
    \multicolumn{1}{c}{} &  &  &  &  & {$1.25\uparrow$} & {(m)$\downarrow$} & {(m) $\downarrow$} & {(deg) $\downarrow$} \\ \hline
    DPSNet$\dagger$~\cite{im2019dpsnet} & \checkmark & 0.199 & 0.142 & 0.438 & 0.710 & - & - & - \\
    CNMNet$\dagger$~\cite{long2020occlusion} & \checkmark & 0.161 & 0.083 & 0.361 & 0.766 & - & - & - \\
    NeuralRecon$\dagger$~\cite{sun2021neuralrecon} & \checkmark & 0.155 & 0.104 & 0.347 & 0.820 & - & - & - \\ \hline
    DeepV2D~\cite{teed2018deepv2d} &  & 0.162 & 0.092 & 0.380 & 0.767 & 0.471 & 1.018 & 60.979 \\
    DROID-SLAM~\cite{teed2021droid} & \multicolumn{1}{l}{} & 0.209 & 0.132 & 0.462 & 0.665 & 0.463 & 0.928 & 40.143 \\ \hline
    Deep3D~\cite{lee20213d} &  & 0.172 & 0.105 & 0.406 & 0.748 & 0.310 & 0.306 & 8.665 \\
    CVD2~\cite{kopf2021robust} &  & 0.154 & 0.085 & 0.379 & 0.795 & 0.375 & 0.517 & 31.102 \\
    Ours (GCVD) &  & \textbf{0.124} & \textbf{0.054} & \textbf{0.307} & \textbf{0.858} & \textbf{0.249} & \textbf{0.257} & \textbf{8.155} \\ \hline
    \end{tabular}
    \caption{\textbf{Quantitative evaluations of depth and pose on 7-Scenes dataset~\cite{shotton2013scene}.} The standard depth evaluation measures per-frame errors and accuracy metrics. The pose evaluation computes the per-sequence pose errors. 
    $\dagger$We also refer to the approaches of multi-view stereo and 3D reconstruction with known camera pose and compare with the results reported in NeuralRecon~\cite{sun2021neuralrecon}.}

    \label{tab:7scenes}
    \vspace{0cm}\vspace{-0.5cm}
\end{table*}

%\subsubsection{Objectives}\label{sec:objectives}
\noindent\textbf{Objectives.}
%As proposed by prior approaches~\cite{lee20213d}, the 
The proposed objectives include inter-frame constraints for geometrical consistency and intra-frame regularization. The inter-frame objectives ensure consistent estimations between two views. % by using SfM. The self-supervised %framework 
The test-time learner exploits the point transformation between $(I_a, I_b)$ via 3D projection as formulated below:
\begin{equation}\label{eq:3d-projection}
    \tilde{x}_b \sim KP_bP_a^{-1}D_a(x_a)K^{-1}\tilde{x}_a,
\end{equation}
 where $\tilde{x}_a$ and $\tilde{x}_b$ denote the homogeneous form of a pixel in $I_a$ and $I_b$, respectively. $K$ stands for the camera intrinsics. Accordingly, the rigid flow $F^b_a=x_b-x_a$ is used to realize the inter-frame constraints in the following three aspects.

%\noindent \textbf{Photometric loss} 
The photometric loss computes the appearance bias % difference 
between $I_a$ and the synthesized $I^b_a$ %via coordinate alignment 
(warped using $F^b_a$) %and $I_b$ 
by $L_1$ and structure dissimilarity~\cite{wang2004image} losses:
\begin{equation}
    \mathcal{L}^{photo}_{a,b} = \frac{1}{|V^b_a|}\sum_{x\in V^b_a}\|I^b_a - I_a\|_1 + DSSIM(I^b_a,I_a),
\end{equation}
where $V^b_a$ denotes the valid points projected from $I_b$ onto the image plane of $I_a$ and excluding likely-dynamic pixels by $\Hat{M}_a$.

%\noindent \textbf{Optical flow loss} 
The optical flow loss measures the displacement error in the image space. Hence, the flow $\Hat{F}$ generated in the pre-processing is used as the supervision for the rigid flow $F$. We examine the forward-backward consistency
between the pre-processing flows $\Hat{F}^b_a$ and $\Hat{F}^a_b$ to form a binary mask $\tilde{V}^b_a$, and let $\Hat{V}^b_a = \tilde{V}^b_a \cap V^b_a$.
\begin{equation}
    \mathcal{L}^{flow}_{a,b} = \frac{1}{|\Hat{V}^b_a|}\sum_{x\in \Hat{V}^b_a}\|F^b_a - \Hat{F}^b_a\|_1.
\end{equation}
%where $\Hat{V}^b_a$ denotes the binary mask further considering the forward-backward consistency between $\Hat{F}^b_a$ and $\Hat{F}^a_b$ in $V^b_a$.

%\noindent \textbf{Depth consistency loss} 
The depth consistency loss $\mathcal{L}^{const}$% proposed by Bian~\etal
~\cite{bian2019unsupervised} assesses the inconsistency between individually estimated depth $D_a$ and $D_b$. %The depth consistency loss  is fomulated as:
\begin{equation}
    \mathcal{L}^{const}_{a,b} = \frac{1}{|V_a|}\sum_{x_a\in V_a}\frac{\|D^b_a - D_a\|_1}{D^b_a + D_a},
\end{equation}
where $D^b_a$ is the transformed depth of $I_a$ using $D_b$, $P_bP_a^{-1}$.

%\subsubsection{Intra-frame objectives}\label{intra-frame-loss}
%\noindent\textbf{Depth gradient loss} 
Apart from inter-frame losses, the intra-frame objectives are utilized to regularize each depth $D_t$, including the dynamic areas. 
We conduct the depth gradient loss $\mathcal{L}^{grad}_t$ to preserve the depth prior from pre-computed MiDaS depth $\Hat{D}_t$. %We find that 
The optimization %joint optimization of depth and pose 
can refine the bias of initial single depth. % while is also prone to lose the prior from initial depth 
Thus, the single depth $\Hat{D}_t$ obtained in the pre-processing is exploited to provide the supervision of depth edge %detail 
(Fig.~\ref{fig:loss_depth_grad}). 
Let $\nabla D^s_t(x)$ denote the 2D gradient vector of pixel $x$ in the downsampled depth map $D_t^s, s \in \{0, 1, 2\}$.
We measure the orientation difference of depth gradients to avoid scale difference between $\Hat{D}_t$ and $D_t$.
\begin{equation}
    \mathcal{L}^{grad}_t = \sum_{s}\sum_{x} \left(1 - \frac{\nabla D^s_t(x) \cdot \nabla\Hat{D}^s_t(x)}{\|\nabla D^s_t(x)\|_2\|\nabla\Hat{D}^s_t(x)\|_2}\right)^2.
\end{equation}

%Note that we use  MiDaS~\cite{Ranftl2020} depth as the target $\Hat{D}_t$ in the experiments. Yet it can be changed to any other improved single depth approaches~\cite{miangoleh2021boosting} to get even better depth detail.

%\noindent \textbf{Deformation regularization} 
The regularization of deformation proposed by Kopf~\etal\cite{kopf2021robust} maintains the spatial smoothness of learnable mesh for a flexible deformation. Likewise, we use the regularization loss $\mathcal{L}^{deform}_t$ %adopts the adaptive weights 
to encourage %more 
smoothness in dynamic area $1-\Hat{M}_t$.
%\noindent \textbf{Final objective}
Finally, the total loss $\mathcal{L}$ of a pair of $(I_a, I_b)$ is %formulated 
conducted as:
%\begin{equation}
\begin{align*}
\mathcal{L}=&\lambda^{photo}\mathcal{L}^{photo}+\lambda^{flow}\mathcal{L}^{flow}+\lambda^{const}\mathcal{L}^{const}\\&+\lambda^{grad}\mathcal{L}^{grad}+\lambda^{deform}\mathcal{L}^{deform},
\end{align*}
%\end{equation}
where the inter-frame objectives compute the bidirectional losses
%(\ie, $\mathcal{L}^{flow}=\mathcal{L}^{flow}_{a,b}+\mathcal{L}^{flow}_{b,a}$)
and the intra-frame objectives sum up the losses of individual frames.
%(\ie, $\mathcal{L}^{grad}=\mathcal{L}^{grad}_{a} + \mathcal{L}^{grad}_{b}$).
The weights $\lambda^{photo}$, $\lambda^{flow}$, $\lambda^{const}$, $\lambda^{grad}$, $\lambda^{deform}$ are set as $1,10,0.5,0.1,0.5$, respectively. Note that $\mathcal{L}^{flow}$ is used only when $I_a$ and $I_b$ are adjacent.

\begin{figure*}[t]
    \centering
    \includegraphics[width=\linewidth]{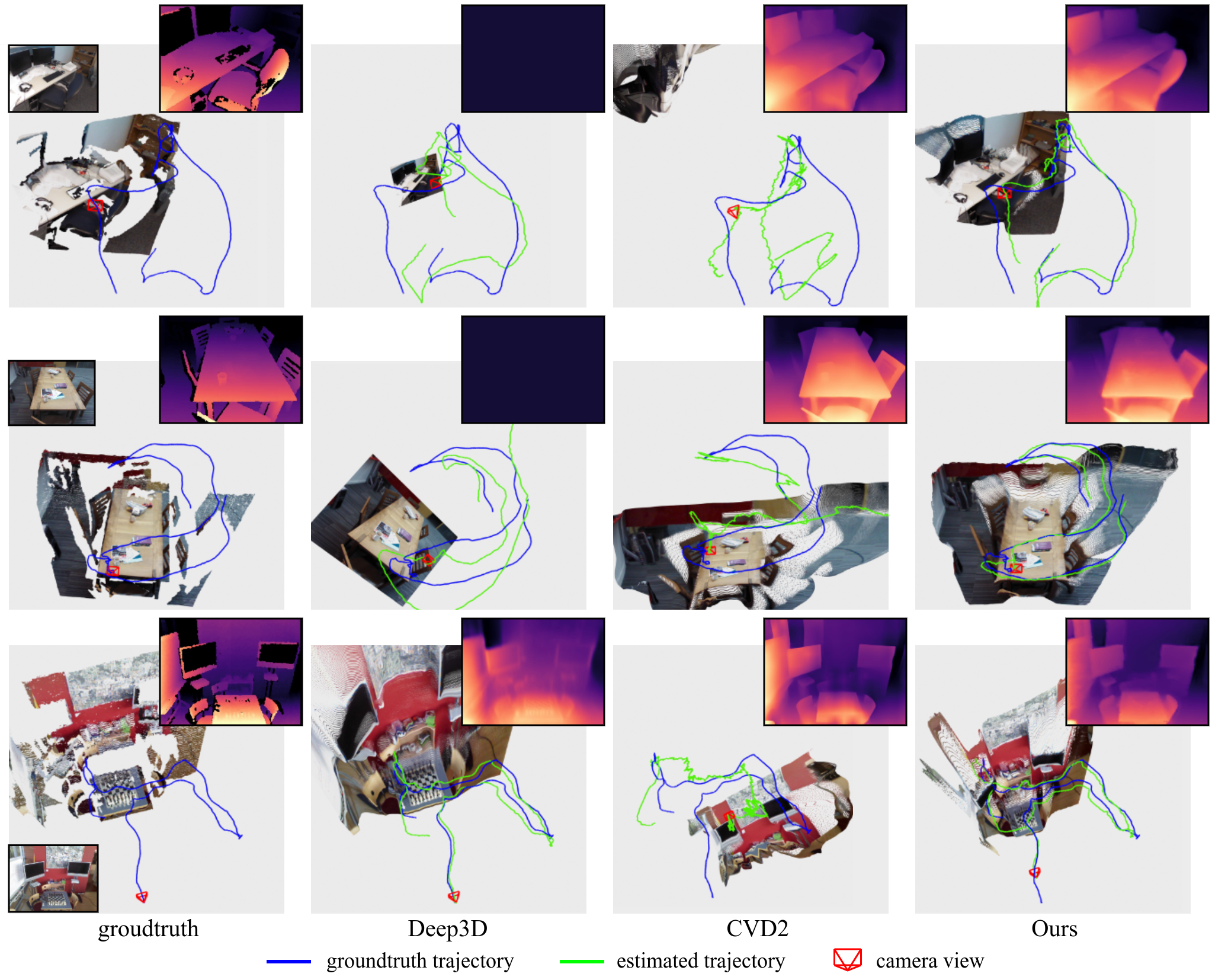}
    \vspace{0cm}\vspace{-0.6cm}
    \caption{\textbf{Visual comparisons with state-of-the-art on 7-Scenes~\cite{shotton2013scene}.} %The green and blue trajectories represent the aligned estimation and groundrtuth camera poses, respectively. 
    The 3D point cloud is back-projected from a view with the estimated depth. 
    Deep3D~\cite{lee20213d} produces in weak depth results and even collapse (constant value).
    CVD2~\cite{kopf2021robust} results in poor pose performance (observed via the bad alignment of the green and blue trajectories)  % chusong:
    despite the plausible depth.
    Our method shows the most favorable performance on both depth and pose. Zoom in for more detailed visualization is suggested.}
    \vspace{0cm}\vspace{-0.5cm}
    \label{fig:7scenes}
\end{figure*}

\subsection{Implementation Detail}
The approach is realized in PyTorch with Adam and g2o library~\cite{kummerle2011g}. 
The resolution of depth and deformation mesh are 384 and 17, respectively, following CVD2~\cite{kopf2021robust} for the longer side of frame. %Therefore, the framework 
%It optimizes 40 frames per mini-batch with a 
RTX3090 GPU is used on the mini-batch size 40. % simultaneously. %at a time. 
We run the optimizations of sequential keyframes, non-sequential keyframes, %pairs, 
and non-keyframes with 300, 100 and 100 iterations, respectively.
We further perform flow-guided depth filter like CVD2~\cite{kopf2021robust} as post-processing. % to refine the depth details. 
The optical flow~\cite{teed2020raft} is used. More details are given in the appendix. %supplementary material.

\section{Experiments}

We compare the proposed method with the SOTA test-time-training methods, CVD2~\cite{kopf2021robust} and Deep3D~\cite{lee20213d}. %Both %CVD2 and Deep3D estimate depth and pose through test-time optimization without COLMAP~\cite{schonberger2016structure} %in advance beforehand. 
CVD2 jointly optimizes pose and learnable deformation %for
from initial MiDaS~\cite{Ranftl2020} depths. Deep3D takes DepthNet and PoseNet~\cite{godard2019digging} and learns from ImageNet pretrained weight to acquire depth and pose. For fair comparisons, we assume an ideal camera intrinsic and re-implement Deep3D with the same resolution of depth, %as well as the same
optical flow estimation~\cite{teed2020raft} and static masks as CVD2 and ours. 
In addition, we compare our approach with the SOTA supervised SLAM systems~\cite{teed2018deepv2d,teed2021droid}, where the SLAM mode of DeepV2D~\cite{teed2018deepv2d} performs %local 
pose optimization in a tracking window and DROID-SLAM~\cite{teed2021droid} utilizes dense bundle adjustment to achieve global consistency.
%which are trained in the other labeled datasets~\cite{Silberman:ECCV12,dai2017scannet,wang2020tartanair}.}

\noindent \textbf{Datasets.}
In contrast to CVD2 conducting %quantitative 
evaluations on synthetic video clips (with each only 50-frames long) of Sintel dataset~\cite{Wulff:ECCVws:2012},
We conduct the experiments on %two   chusong:
long sequences (\textbf{500 to 3000 frames}) of 
\textbf{real-world datasets}. %which provide groundtruth depth and camera trajectories.

\noindent\textit{$\bullet$~7-Scenes RGB-D dataset}~\cite{shotton2013scene} has 46 sequences with either 500 or 1000 frames. The indoor scenes are grabbed with a Kinect camera at size 640$\times$480. % resolution.% The groundtruth camera poses are measured by KinectFusion~\cite{newcombe2011kinectfusion}.

\noindent\textit{$\bullet$~TUM RGB-D dataset}~\cite{sturm12iros} is gathered by a handheld Kinect camera %with groudtruth trajectories from motion capture systems 
with more demanding cases such as texture-less area and abrupt motions. %Several 
Seven representative sequences (613$\sim$2965 frames) in TUM RGB-D are used for evaluation.

\noindent\textit{$\bullet$~EuRoC dataset}~\cite{Burri25012016} has 11 sequences (1710$\sim$3682 frames) filmed by a micro aerial vehicle. We demonstrate the comparisons in our appendix. %supplementary material.
%We also demonstrate the comparisons on more challenging \textbf{EuRoC dataset} (1710$\sim$3682 frames)~\cite{Burri25012016} in our supplementary material. 

\noindent\textbf{Evaluation metrics.}
We follow the standard depth evaluation~\cite{eigen2014depth} to align the scales between the estimated and groundtruth depths by median scaling. 
%On the other hand, the 
The pose evaluation uses the %evaluation protocol 
metric of visual odometry~\cite{sturm12iros,Zhang18iros}, including absolute trajectory error (ATE) and relative pose error (RPE) with 7-DoF alignment.

\begin{figure*}[h]
    \centering
    \includegraphics[width=\linewidth]{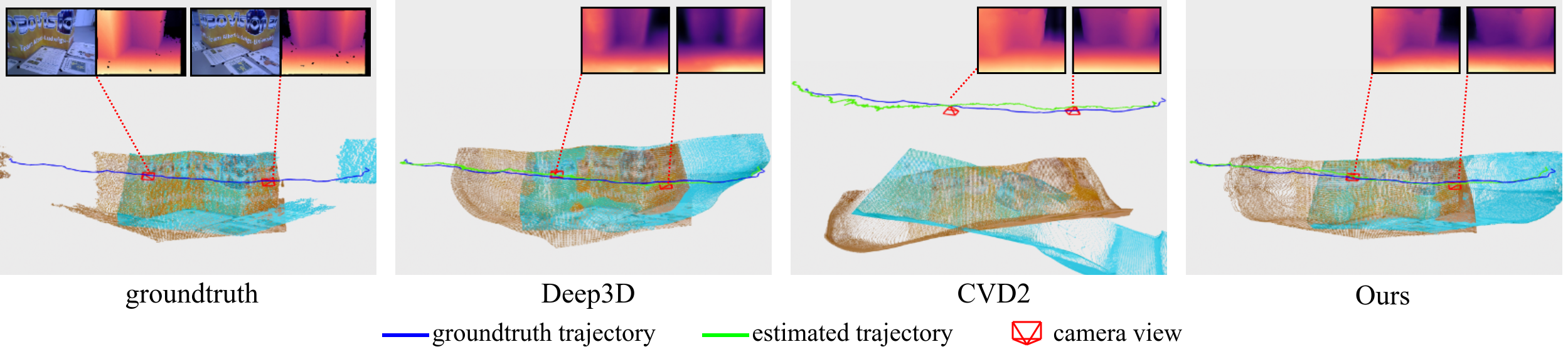}
    \vspace{0cm}\vspace{-0.3cm}
    \caption{\textbf{Visual comparisons of geometric consistency.}
    Visualizing the geometric consistency via the point clouds %alignments 
    between two distant views (200-frames interval). As %demonstrated 
    shown in the groundtruth, the back-projected point clouds (brown and cyan) from two views are supposed to align in the co-visible area. Our GCVD and Deep3D~\cite{lee20213d} maintain better geometric consistency and show promising alignments of point clouds, while CVD2~\cite{kopf2021robust} results in severe misalignment between the distant views.
    % \textcolor{red}{(?? How to observe the "geometric consistency"?}
    }
    \label{fig:tum}
\end{figure*}

\begin{table*}[t]
    \centering
    %\scriptsize
    \begin{tabular}{l|C{1.3cm}C{1.25cm}C{1.25cm}C{1.45cm}|C{1.25cm}C{1.25cm}C{1.45cm}}
    \hline
    \multirow{2}{*}{Sequence} & \multicolumn{4}{c|}{Pose Error (ATE in meters) $\downarrow$} & \multicolumn{3}{c}{Depth Error (Abs Rel) $\downarrow$} \\ \cline{2-8} 
     & \multicolumn{1}{c|}{COLMAP} & Deep3D & CVD2 & GCVD & Deep3D & CVD2 & GCVD \\ \hline
    \texttt{fr1\_desk} & \multicolumn{1}{c|}{0.019} & 0.580 & 0.273 & {0.229} & 0.1940 & 0.1090 & {0.0940} \\
    \texttt{fr1\_desk2} & \multicolumn{1}{c|}{0.027} & 0.611 & 0.314 & {0.156} & 0.2282 & {0.1139} & 0.1305 \\
    \texttt{fr2\_desk} & \multicolumn{1}{c|}{0.818} & {0.260} & 0.613 & 0.422 & {0.0973} & 0.1544 & 0.1130 \\
    \texttt{fr3\_cabinet} & \multicolumn{1}{c|}{failed} & 1.225 & {0.444} & 0.850 & 0.4613 & {0.1236} & 0.1832 \\
    \texttt{fr3\_nstr\_tex\_near\_loop} & \multicolumn{1}{c|}{0.015} & 0.694 & 0.491 & {0.399} & 0.2437 & 0.0615 & {0.0352} \\
    \texttt{fr3\_sitting\_static} & \multicolumn{1}{c|}{0.033} & 0.006 & 0.029 & {0.006} & 0.2368 & 0.1260 & {0.1243} \\
    \texttt{fr3\_str\_tex\_far} & \multicolumn{1}{c|}{0.008} & {0.089} & 0.169 & 0.131 & {0.0511} & 0.0742 & 0.0810 \\ \hline
    Mean & \multicolumn{1}{c|}{0.153} & 0.495 & 0.333 & \textbf{0.313} & 0.2160 & 0.1089 & \textbf{0.1087} \\ \hline
    \end{tabular}
    \caption{\textbf{Quantitative evaluation on TUM RGB-D~\cite{sturm12iros}}. We present the depth and pose errors of each sequence and compare with %DeepV2D~\cite{teed2018deepv2d}, 
    Deep3D~\cite{lee20213d}, and CVD2~\cite{kopf2021robust}. We additionally provide the pose results of COLMAP~\cite{schonberger2016structure} as reference.}
    \label{tab:tum}
    %\vspace{0cm}\vspace{-0.1cm}
\end{table*}

\subsection{Evaluation on 7-Scenes~\cite{shotton2013scene}}
Table~\ref{tab:7scenes} shows the quantitative comparisons of our approach to the SOTA methods. 
Although DeepV2D~\cite{teed2018deepv2d} and DROID-SLAM~\cite{teed2021droid} utilize pose graph for the pose refinement and bundle adjustment in testing, they are restricted by the generalization capability of supervised learning in 
different datasets.
Test-time training approach Deep3D~\cite{lee20213d} shows the second best pose performance; however, it results in worse depth estimation. %without MiDaS~\cite{Ranftl2020} as the depth prior.   chusong:
CVD2~\cite{kopf2021robust} maintains more depth priors from MiDaS while the bias in the depth prior leads in poor pose estimation. 
Besides, we compare our GCVD with other depth-estimation approaches with known camera pose. The quantitative depth scores of DPSNet~\cite{im2019dpsnet}, CNMNet~\cite{long2020occlusion}, and NeuralRecon~\cite{sun2021neuralrecon} are provided by~\cite{sun2021neuralrecon}. Similar to DeepV2D and DROID-SLAM, the supervised methods with known pose may still suffer from the domain discrepancy between training and test data. In contrast, the proposed 
GCVD achieves globally consistent optimization on test data and demonstrates the best performance on both standard depth and pose metrics.

We conduct visual comparisons by displaying the back-projected point clouds and the pose estimation with the groundtruth trajectory by 7-DoF alignment~\cite{Zhang18iros}. 
As shown in Fig.~\ref{fig:7scenes}, although Deep3D demonstrates some promising camera estimations, it is prone to collapse on the depth estimation. 
CVD2 provides plausible depths with the aids of MiDaS while yields weak pose results due to the lack of global consistency. 
Again, our method %outperforms Deep3D and CVD 
reveals better results in 3D visualization.

\subsection{Evaluation on TUM-RGBD~\cite{sturm12iros}}
In this experiment, we compare our GCVD with the test-time training approaches (CVD2 and Deep3D) and also COLMAP which is a traditional representative solution having maintaining global consistency.
Table~\ref{tab:tum} provides the per-sequence quantitative errors. 
We only present the pose results of COLMAP %\cite{schonberger2016structure} 
since its dense depths via multi-view stereo still contain holes and noises. 
Although COLMAP shows superior pose results, it completely fails (collapses) in a sequence (marked as red in Table~\ref{tab:tum}) due to the fragile sparse reconstruction.
%shows superior pose results, while it is fragile due to the sparse reconstruction and thus fails in a sequence. \textcolor{blue}{Besides, we do not consider the depth of COLMAP since it requires even more expensive overhead for multi-view stereo and still yielding holes and noises.} 
As for test-time training, our method attains the lowest depth and pose errors in overall. We handle depth prior properly with the learnable networks to facilitate pose learning. Thus, our approach shows better pose estimation on most sequences compared with CVD2, %which suffers from the bias of the fixed MiDaS depth.
which accumulates more pose errors for long sequences. 
Our GCVD can refine the long pose estimation for maintaining global consistency.
On the other hand, we find the weak pose performance of some sequences affected by the unreliable pre-computed optical flow, % that 
which will be discussed in the limitations of our approach in the appendix. %supplementary material.
%Overall, our method performs the lowest depth and pose errors in contrast to learning-based CVD2 and Deep3D. \textcolor{blue}{Yet we found that the pose estimation is mainly restricted by the poor performance of the precomputed optical flow in some sequences due to large texture-less regions and motion blur.} For depth scores, our method and CVD2 still produce superior results due to the strength of MiDaS, whereas Deep3D results in poor depths.

Besides, we compare the geometric consistency by visualizing the alignment of point clouds from different views.
The two distant frames viewing the common scenes are back-projected the 3D point clouds. Hence, the geometric inconsistency can be seen by the misalignment between the two point clouds in the
co-visible area. 
As shown in Fig.~\ref{fig:tum}, Deep3D and ours show similar point cloud alignments to the groundtruth's because both approaches have the depth fine-tuning mechanisms. % by leveraging CNN-based optimizations to refine the geometric consistency.
Nevertheless, the accumulated depth bias leads CVD2 to yield poor pose estimation and large misalignment between the views. % of about 200-frames interval.
%colors in order to visualize the consistency via point cloud alignment.} As can be seen, %Deep3D provides plausible consistency with small difference of alignment while provides sub-optimal depth estimation in detail. Nevertheless, 
%\textcolor{blue}{even though CVD2 shows promising depth visualization in each view, the poor temporal consistency affected by the bias of fixed MiDas depth can be seen from the large difference of point cloud alignments.} \textcolor{red}{In contrast, Deep3D and Our results leverage CNN-based optimization framework to refine geometrically consistent 3D estimations among distant views and better depth estimation than Deep3D.}

\subsection{Ablation studies}
We conduct ablation studies on 7-Scenes dataset~\cite{shotton2013scene} in Table~\ref{tab:ablation}. For the depth component, the inserted layer normalization stabilizes the depth scale of MiDaS~\cite{Ranftl2020} network and hence facilitates the depth and pose performance by 35\% and 40\%, respectively. The flexible deformation~\cite{kopf2021robust} regulates the spatial misalignments of each depth to improve depth estimation by 4\%. 
Besides, the depth gradient loss %is used to retain 
retains the initial depth priors during refining the depth bias. 
The well-handled depth prior can encourage a better joint optimization, thus reducing the pose errors by 15\% (0.327 to 0.277). Moreover, the method without the keyframe strategy raises 11\% pose error (0.277 to 0.308) due to the lack of proper camera baselines for initial optimization. Finally, the pose graph optimization for global consistency further cuts down the pose error by 10\% (0.277 to 0.249).

\begin{table}[t]
    \centering
    \small
    \begin{tabular}{C{0.5cm}C{0.85cm}C{1cm}C{0.7cm}C{0.6cm}C{0.8cm}C{0.8cm}}
    \hline
    \multicolumn{5}{c}{Ablation settings} & \multicolumn{2}{c}{Errors $\downarrow$} \\
    \multirow{2}{*}{KF} & layer & mesh & grad. & \multirow{2}{*}{PGO} & \multicolumn{1}{c}{Depth} & \multicolumn{1}{c}{Pose} \\
     & norm & deform. & loss &  & \multicolumn{1}{c}{AbsRel} & \multicolumn{1}{c}{ATE} \\ \hline
    \checkmark &  &  &  &  & 0.208 & 0.555 \\
    \checkmark & \checkmark &  &  &  & 0.137 & 0.332 \\
    \checkmark & \checkmark & \checkmark &  &  & 0.132 & 0.327 \\
    \checkmark & \checkmark & \checkmark & \checkmark &  & 0.125 & 0.277 \\
     & \checkmark & \checkmark & \checkmark &  & 0.127 & 0.308 \\
    \checkmark & \checkmark & \checkmark & \checkmark & \checkmark & \textbf{0.124} & \textbf{0.249} \\ \hline
    \end{tabular}
    \caption{\textbf{Ablation studies on 7-Scenes~\cite{shotton2013scene}.} We validate the effectiveness of the keyframe strategy (KF), the added layer normalization and mesh deformation in the depth component, depth gradient loss, and pose graph optimization (PGO) in our pipeline.}
    \label{tab:ablation}
\end{table}

\subsection{Runtime comparisons}
We compare the runtime of our %method 
GCVD with the test-time-training methods. We select five long videos %from EuRoC dataset~\cite{Burri25012016} 
and extract the first $n$ frames of the videos to compose different lengths of sequences for runtime evaluation. 
%Therefore, the runtime 
The execution times of the videos are measured on an i7-11700K with a RTX3090 GPU. 
We present the averaged per-frame %runtime 
time in Table~\ref{tab:time}. 
Note that we do not compare with COLMAP which requires extremely expensive time (\eg, $\sim$56 secs for each frame on a 2000-frame video). CVD2 shows about 6 to 9 times slower than our method due to the preparation of multiple pairs of optical flow and the traditional %optimization
optimizer with CPU. 
Deep3D provides strong efficiency in long videos; %, yet 
however, it tends to yield drifts and collapse in depth estimation. 
In contrast, our method %has the fastest speed 
performs fastest in short videos by learning few keyframes first then optimizing the non-keyframes with fewer iterations. 
For long videos, our GCVD is slightly slower than Deep3D due to keyframe association on more keyframes for global consistency.
%tackling global consistency. %the geometric verification in 
%Our GCVD requires a bit more time on keyframe association for more keyframes in long sequences.} %More analyses are presented in the supplementary material.
% chusong:??

\begin{table}[t]
    \centering
    \footnotesize
    \begin{tabular}{L{1.2cm}R{0.9cm}R{0.9cm}R{0.9cm}R{0.9cm}R{0.9cm}}
    \hline
    \# frames & 50 & 200 & 500 & 1000 & 2000 \\ \hline
    CVD2 & 13.90 & 14.83 & 17.34 & 21.73 & 26.13 \\
    Deep3D & 4.13 & 3.03 & 2.75 & 2.66 & 2.61 \\
    GCVD & 2.40 & 2.53 & 2.74 & 2.83 & 3.04 \\ \hline
    \end{tabular}
    \caption{\textbf{Runtime comparison.} We compute the per-frame runtime (sec) in different lengths of videos. Our method shows the fastest speed in short videos and achieves a good trade-off between efficiency and global consistency compared with Deep3D.}
    \label{tab:time}
\end{table}

\section{Conclusion}
% yaochih: I am not confident to claim that we are the first of something.
We present GCVD, a %robust 
learning-based method for video depth and pose estimation with global consistency and efficiency. %, both are vital for long videos. To achieve both goals, the 
%The proposed method 
To our knowledge, this is the first study %introduces a keyframe-based pose graph with global pose optimization into the CNN-based optimization framework \textcolor{black}{for simultaneous depth and pose estimation}.
tackling global consistency for test-time training.
Based on the global poses of keyframes from the pose-only bundle adjustment, the deep networks jointly learn keyframe depth refinement and the depth and pose of the remaining frames efficiently.
In addition, our proposed method can better handle single depth prior properly and fine-tune the depth network to alleviate depth bias and achieve robust and consistent 3D estimation.
%The flow-guided keyframes are utilized to construct the pose graph %with 
%from deep network features to detect the %-associated 
%associated edges %and attain robust estimation 
%as initials; thus, the depth and pose of the remaining frames can be estimated efficiently with the pose graph and deep models.
%Besides, our GCVD handles depth priors properly to alleviate the bias of single depth and avoid from collapse.}
%\textcolor{blue}{Besides integrating the advantages of the prior learning-based approaches, the proposed pipeline extracts keyframes with optical flow as guidance; therefore, the CNN-based optimization framework is able to perform robust estimation with proper camera movement among keyframes as initials. Furthermore, we construct a keyframe-based pose graph with CNN-feature-associated loop edges so as to attain globally-consistent estimation.} 
Experimental results show that GCVD outperforms the state-of-the-art approaches on both depth and pose evaluation. 
Moreover, GCVD achieves %strong 
high efficiency %with global consistency.
by %using the strategy of tilizing the keyframes and 
keeping the scene knowledge in network weights to boost the optimization of next fragment of frames.
%The discussions of limitations 
%Further discussions on the limitations %of the proposed method 
%and our future directions are %presented 
%given in %our 
%the supplementary material. 
We will release our codes to public.
In contrast to COLMAP that uses traditional techniques, our GCVD is a fundamental deep-learning tool for the offline-video SfM. 
%The future directions include handling varying camera intrinsics for zooming videos to further improve the capability.

\clearpage

\appendix

\section{Appendix}

We present GCVD, a test-time training method for video-based 3D estimation based on offline videos.
There are still few test-time training studies for video-based SfM.
Existing approaches are, however, robust to only local consistency for video depth estimation.
Our approach is the first global-consistency solution to this direction.
It needs only light-weight pose-only bundle adjustment as initial, and then takes advantage of neural-networks learning for global optimization of poses and depths simultaneously.
Our GCVD can serve as a generally useful tool for offline video-based SfM (like the renowned tool COLMAP using traditional approach), where it can provide dense 3D estimations instead of fragile or sparse 3D outputs for challenging conditions such as homogeneous areas and motion blurs.
Compared to the representative test-time-training approaches (such as CVD2~\cite{kopf2021robust}), our approach can handle long videos at reasonable runtime.
Unlike the approach of~\cite{kopf2021robust} that only validates the performance using 50-frames video clips, we validate the performance for 500$\sim$3682 frames, which considerably boosts the validation to practically useful situations.
%The core codes are provided for review but not runnable. 
%The complete implementation will be released to publish later.
%The codes of our implementation are provided in the supplementary file.

In this appendix, we present additional details to complement our main paper, including \textit{implementation details}, 
\textit{comparisons with the state-of-the-art SLAM approaches},
\textit{evaluation on \emph{EuRoC dataset}}~\cite{Burri25012016}, \textit{runtime analysis},
and \textit{limitations of our GCVD}.
%and detailed runtime analysis (Sec.~\ref{sec:runtimeanalysis}).
%We also encourage readers to see visual comparison in the supplementary video. 

%\textcolor{blue}{\href{https://drive.google.com/file/d/1ev9BEcZCI7cQFQsHZSZVfQj5Pu6ofY50/view?usp=sharing}{yaochih: the video file is attached here temporarily}}.

%\begin{figure}
%    \begin{center}
%	\includegraphics[width=0.7\linewidth]{figures/loss_pe.png}
%	\end{center}
%	\vspace{-0.3cm}
%	\caption{\textbf{The effectiveness of positional encoding in the pose module.} The convergence curves of keyframe and non-keyframes with positional encoding (\textit{p.e.}) show better convergence speeds and even achieve the lower loss in non-keyframe optimization.}
%	\label{fig:loss_pe}
%\end{figure}

\subsection{Implementation Detail}\label{sec:implementdetail}

\begin{figure}[]
    \centering
    \includegraphics[width=\linewidth]{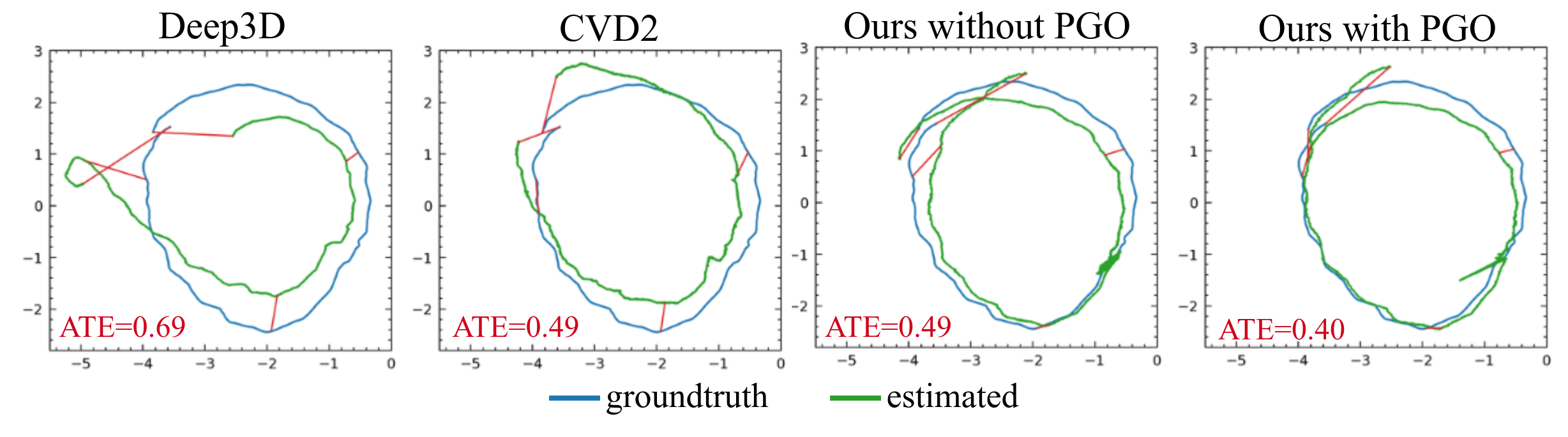}
    \caption{{\textbf{Visualization of global consistency.} Our method reduces 18\% pose error by using pose graph optimization (PGO) on the sequence {\fontfamily{qcr}\selectfont fr3\_nstr\_tex\_near\_loop} in TUM RGB-D dataset~\cite{sturm12iros}.}}
    \label{fig:loop}
\end{figure}

\noindent\textbf{Keyframe-based pose graph optimization.}
We perform pose graph optimization with g2o~\cite{kummerle2011g} for globally consistent pose estimation. 
The edges of the pose graph include \emph{sequential} and \emph{non-sequential} %co-visible
edges.
Each sequential edge $e_{i,i\pm\tau}, \tau\in\{1,2,4,8\}$ of the pose graph connects the keyframe pair $(\kappa_i, \kappa_{i\pm\tau})$ with the relative pose $P^{-1}_{\kappa_i}P_{\kappa_i\pm\tau}$ and the weight matrix $diag(\frac{1}{\tau})$ to ensure temporal coherence in a local range. 
The sequential edges contain relatively nearby views determined by the optical-flows.
However, in addition to the sequential edges, there could be farther views which share co-visible scenes.
Hence, we also establish
%On the other hand, 
the non-sequential edges via keyframe association, which connects the keyframe pairs with the index differences exceeding $\alpha=\max(\tau)$ %, 
but %each pair shares 
sharing a co-visible scene to enhance global consistency.
Similarly, the non-sequential co-visible edges are constructed with the optimized relative poses and identity weight matrices. %Therefore, the 
The pose graph optimization is conducted with at most 100 iterations to obtain global pose estimation $P_{\kappa_{1..k}}$.
Afterward, the depth and pose of the remaining frames and the depth of keyframes are estimated simultaneously with the frozen keyframe poses to maintain global consistency.
Figure~\ref{fig:loop} demonstrates the effectiveness of the pose graph optimization for achieving the global consistency.

\noindent\textbf{Detail of joint depth and pose optimization.}
The test-time optimization framework is implemented in PyTorh with Adam (with $\beta_1=0.9$, $\beta_2=0.999$). 
The learning rates for sequential keyframes, non-sequential keyframes, and non-keyframes are $2\times10^{-4}$, $5\times10^{-5}$, and $1\times10^{-4}$, respectively. 
To speed up the optimization, we compute the loss with a quarter scale of depth estimation in the keyframe optimization. Thus, the depth of keyframes will be further refined in the non-keyframe optimization with original scale (\ie, 384 for the longer side of the frame).

\noindent\textbf{Post-processing.}
We follow the flow-guided depth filter proposed in CVD2~\cite{kopf2021robust} to further enhance the temporal consistency of edge details in depth maps. The final filtered depth $\tilde{D}_t$ acquires the depth details from neighboring depths $D_{t-\Omega..t+\Omega}$ ($\Omega=4$) with the chaining optical flow $\tilde{F}$.
\begin{equation}
    \tilde{D}_t = \sum_{i=t-\Omega}^{t+\Omega}\omega_{i\to t}D_{i\to t},
\end{equation}
where $D_{i\to t}$ is the projected depth of $I_t$ by transforming $D_i$ with $P_t$, $P_{i}$ and the chaining flow $\tilde{F}_{i\to t}$ to align the pixel coordinate. The maximum span $\Omega$ is set as 4 by default, and the weight term $\omega_{i\to t}$ considers both the depth reprojection error and the forward-backward inconsistent error of the chaining flow as follows:
\begin{equation}
    \omega_{i\to t}=\exp\left(-\gamma_1\frac{\max(D_t, D_{i\to t})}{\min(D_t, D_{i\to t})} -\gamma_2 \tilde{F}^{diff}_{i\to t}\right),
\end{equation}
where $\tilde{F}^{diff}_{i\to t}$ denotes the forward-backward inconsistency between the chaining flow $\tilde{F}_{i\to t}$ and $\tilde{F}_{t\to i}$. The $\gamma_1=2$ and $\gamma_2=0.1$ balance the strength of the temporal depth filter. In the end, the final consistent depth $\tilde{D}_{1..N}$ and pose $P_{1..N}$ of the entire video $I_{1..N}$ is accomplished.

\begin{table*}[t]
    \centering
    \small
    \caption{\textbf{Comparison with learning-based SLAMs on 7-Scenes~\cite{shotton2013scene} and TUM RGBD dataset~\cite{sturm12iros}.}}
    \begin{tabular}{C{1.7cm}lR{1cm}R{1cm}R{1cm}R{1cm}R{1cm}R{1.2cm}R{1.2cm}}
\hline
\multirow{3}{*}{Dataset} & \multicolumn{1}{c}{\multirow{3}{*}{Method}} & \multicolumn{4}{c}{Depth Metrics} & \multicolumn{3}{c}{Pose Metrics} \\
 & \multicolumn{1}{c}{} & \multirow{2}{*}{\tiny{AbsRel$\downarrow$}} & \multirow{2}{*}{\tiny{SqRel$\downarrow$}} & \multirow{2}{*}{\tiny{RMSE$\downarrow$}} & \tiny{$\delta<$} & \tiny{ATE} & \tiny{RPE Trans} & \tiny{RPE Rot} \\
 & \multicolumn{1}{c}{} &  &  &  & \tiny{$1.25\uparrow$} & \tiny{(m)$\downarrow$} & \tiny{(m) $\downarrow$} & \tiny{(deg) $\downarrow$} \\ \hline
\multirow{3}{*}{\begin{tabular}[c]{@{}c@{}}7-Scenes\\ \end{tabular}} & DeepV2D~\cite{teed2018deepv2d} & 0.162 & 0.092 & 0.380 & 0.767 & 0.471 & 1.018 & 60.979 \\
 & DROID-SLAM~\cite{teed2021droid} & 0.209 & 0.132 & 0.462 & 0.665 & 0.463 & 0.928 & 40.143 \\
 & Ours & \textbf{0.124} & \textbf{0.054} & \textbf{0.307} & \textbf{0.858} & \textbf{0.249} & \textbf{0.257} & \textbf{8.155} \\ \hline
\multirow{3}{*}{\begin{tabular}[c]{@{}c@{}}TUM\\ RGBD\\ \end{tabular}} & DeepV2D~\cite{teed2018deepv2d} & 0.166 & 0.153 & 0.648 & 0.745 & 0.460 & 1.360 & 60.479 \\
 & DROID-SLAM~\cite{teed2021droid} & 0.214 & 0.211 & 0.778 & 0.639 & \textbf{0.013} & 1.327 & 50.794 \\
 & Ours & \textbf{0.109} & \textbf{0.077} & \textbf{0.461} & \textbf{0.858} & 0.313 & \textbf{0.277} & \textbf{15.919} \\ \hline
\end{tabular}
    \label{tab:learning-slams}
\end{table*}

\subsection{Comparison with SOTA learning-based SLAMs}\label{sec:learning-slam}
Although the SLAM approach reconstructs depth and pose from online (streaming) videos, which is different from our problem setting of video SfM for offline videos, we compare with the state-of-the-art supervised SLAMs~\cite{teed2018deepv2d,teed2021droid} which tackle global consistency.
DeepV2D~\cite{teed2018deepv2d} performs global pose optimization with a tracking window of eight frames. DROID-SLAM~\cite{teed2021droid} utilizes dense and full bundle adjustment to achieve global consistency.
Table~\ref{tab:learning-slams} presents the quantitative comparison on 7-Scenes~\cite{shotton2013scene} and TUM RGBD~\cite{sturm12iros} datasets.
Though DROID-SLAM achieves a superior score on the pose metric of Absolute Trajectory Error (ATE) on TUM RGBD, it shows the deficiency on the other pose metric, Relative Pose Error (RPE), which is used for measuring the drift.
%Furthermore, both 
Both DeepV2D and DROID-SLAM are supervised methods trained on other datasets with groudtruth depth or pose. They suffer from generalization ability due to domain discrepancy and thus result in weak results on 7-Scenes dataset. In contrast, our test-time training approach directly learns on the input test video to %overcome 
address the generalization issue and achieves the best scores on 7-Scenes.

\subsection{Comparison with ORB-SLAM2~\cite{mur2015orb}}\label{sec:orb-slam}
We also compare GCVD with the traditional state-of-the-art SLAM approach ORB-SLAM2~\cite{mur2015orb}, which performs loop closure to retain globally consistent poses and 3D map. 
Table~\ref{tab:tum-traditional-comparison} shows the pose comparisons with traditional COLMAP~\cite{schonberger2016structure} and ORB-SLAM2~\cite{mur2015orb} on TUM RGBD dataset~\cite{sturm12iros}. 
In general, ORB-SLAM2 shows the %best 
most accurate pose results with sparse hand-crafted features. 
Nonetheless, the sparse 3D reconstruction cannot provide complete dense depth for various video processing applications. 
Moreover, COLMAP and ORB-SLAM2 suffer from the robustness issues of the fragile hand-crafted features. 
They failed on reconstruction/tracking on one and two sequences as shown in Table~\ref{tab:tum-traditional-comparison}, respectively. 
On the other hand, the test-time training-based approach can overcome the robustness issue and provide dense depths with dense flow. % while also be restricted by the performance of the flow estimator. 
%In addition, 
Note that our GCVD shows promising pose results close to COLMAP's on average (excluding the failed sequences). 
%We believe that our method is a 
Our GCVD thus provides a fundamental video SfM tool %providing 
on dense depth reconstruction for video processing applications.

\begin{table*}
    \centering
    \small
    \caption{\textbf{Comparison of Absolute Trajectory Error (ATE) with traditional COLMAP~\cite{schonberger2016structure} and ORB-SLAM2~\cite{mur2015orb} on TUM RGBD dataset~\cite{sturm12iros}. We only %compare 
    show the pose results since both COLMAP and ORB-SLAM2 perform 3D reconstruction with sparse point clouds instead of dense depths.}}
    \begin{tabular}{lC{1.5cm}C{1.5cm}C{1.5cm}C{1.5cm}C{1.5cm}}
\hline
 & COLMAP & ORB- & Deep3D & CVD2 & Our \\
\multirow{-2}{*}{Sequence} &  & SLAM2 &  &  & GCVD \\ \hline
\texttt{fr1\_desk} & 0.019 & 0.013 & 0.580 & 0.273 & 0.229 \\
\texttt{fr1\_desk2} & 0.027 & {failed} & 0.611 & 0.314 & 0.156 \\
\texttt{fr2\_desk} & 0.818 & 0.009 & 0.260 & 0.613 & 0.422 \\
\texttt{fr3\_cabinet} & {failed} & {failed} & 1.225 & 0.444 & 0.850 \\
\texttt{fr3\_nstr\_tex\_near\_loop} & 0.015 & 0.010 & 0.694 & 0.491 & 0.399 \\
\texttt{fr3\_sitting\_static} & 0.033 & 0.023 & 0.006 & 0.029 & 0.006 \\
\texttt{fr3\_str\_tex\_far} & 0.008 & 0.009 & 0.089 & 0.169 & 0.131 \\ \hline
Mean &  &  &  &  &  \\
(exclude \texttt{fr1\_desk2},\texttt{fr3\_cabinet}) & \multirow{-2}{*}{0.179} & \multirow{-2}{*}{0.013} & \multirow{-2}{*}{0.326} & \multirow{-2}{*}{0.315} & \multirow{-2}{*}{0.237} \\ \hline
\end{tabular}
    \label{tab:tum-traditional-comparison}
\end{table*}

\begin{figure}
    \centering
    \includegraphics[width=\linewidth]{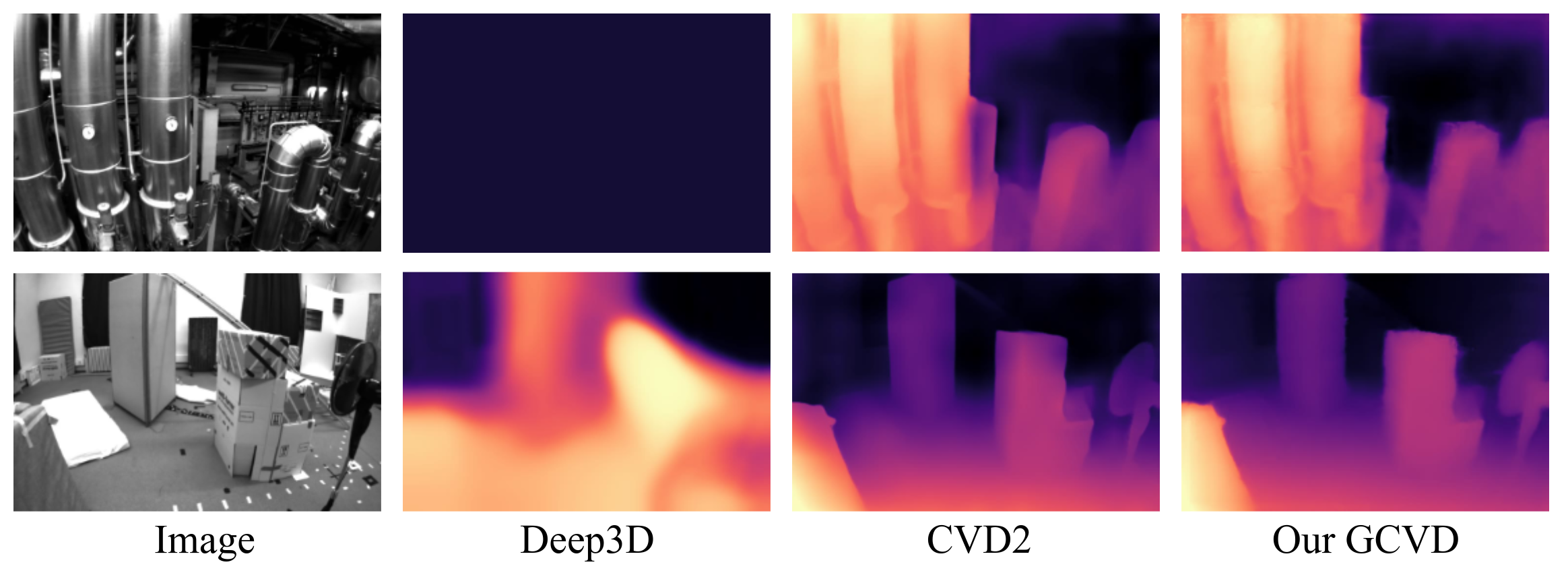}
    \caption{\textbf{Depth visualization on EuRoC dataset~\cite{Burri25012016}.} CVD2~\cite{kopf2021robust} and our method show plausible depths while Deep3D~\cite{lee20213d} tends to collapse with training from ImageNet pretrained weights.}
    \label{fig:visualization-euroc}
\end{figure}

\begin{table*}
    \centering
    \caption{\textbf{Pose evaluation on EuRoC dataset~\cite{Burri25012016}.} The per-sequence absolute trajectory errors (ATE) are reported in meters.}
    \begin{tabular}{lC{0.75cm}C{0.75cm}C{0.75cm}C{0.75cm}C{0.75cm}C{0.75cm}C{0.75cm}C{0.75cm}C{0.75cm}C{0.75cm}C{0.75cm}C{0.75cm}}
\hline
\multirow{2}{*}{Sequence} & MH & MH & MH & MH & MH & V1 & V1 & V1 & V2 & V2 & V2 & \multicolumn{1}{l}{\multirow{2}{*}{\textbf{Mean}}} \\
 & 01 & 02 & 03 & 04 & 05 & 01 & 02 & 03 & 01 & 02 & 03 & \multicolumn{1}{l}{} \\ \hline
Deep3D & 3.24 & 3.82 & 2.82 & 4.26 & 5.06 & 1.53 & 1.57 & 1.32 & 1.97 & 1.87 & 1.73 & 2.65 \\
CVD2 & 1.48 & { 1.32} & 2.63 & 4.20 & { 4.31} & {0.94} & 1.63 & { 1.19} & 1.43 & { 1.56} & 1.74 & 2.04 \\
Our GCVD & { 1.33} & 1.72 & { 1.99} & { 3.78} & 4.59 & 1.11 & { 1.07} & 1.33 & { 0.96} & 1.88 & { 1.46} & \textbf{1.93} \\ \hline
\end{tabular}
    \vspace{0cm}\vspace{0.25cm}
    \label{tab:euroc}
\end{table*}

\subsection{Evaluation on EuRoC~\cite{Burri25012016}}\label{sec:evaleuroc}
The challenging \emph{EuRoC dataset}~\cite{Burri25012016} consists of 11 gray-scale sequences from a stereo camera mounted on a micro aerial vehicle in relatively large indoor environments. 
The groundtruth camera poses are captured by a laser tracker and motion capture system. 
We present the absolute trajectory error (ATE) of each sequence in Table~\ref{tab:euroc} and the qualitative comparison of depth estimation in Figure~\ref{fig:visualization-euroc}. Our method shows the lowest pose error in average. Besides, our GCVD and CVD2 maintain the depth prior of MiDaS while Deep3D %results in
produces sub-optimal depth or collapse. 
%On the other hand, we find that the pose estimation is limited by the performance of optical flow on gray-scale images. 

\begin{table*}[]
    \centering
    \begin{tabular}{lccccccr}
\multicolumn{8}{c}{\textbf{per-frame runtime on a 50-frame video}} \\ \hline
\multicolumn{1}{c|}{} & \multicolumn{2}{c|}{Pre-} & \multicolumn{3}{c|}{Main} & \multicolumn{1}{c|}{Post-} & \multicolumn{1}{c}{\multirow{2}{*}{Sum}} \\
\multicolumn{1}{c|}{} & \multicolumn{2}{c|}{process.} & \multicolumn{3}{c|}{procedure} & \multicolumn{1}{c|}{process.} & \multicolumn{1}{c}{} \\ \hline
\multicolumn{1}{l|}{Deep3D} & \multicolumn{2}{c|}{1.85} & \multicolumn{3}{c|}{2.27} & \multicolumn{1}{c|}{-} & 4.13 \\ \hline
\multicolumn{1}{l|}{CVD2} & \multicolumn{2}{c|}{1.73} & \multicolumn{3}{c|}{7.08} & \multicolumn{1}{c|}{5.09} & 13.90 \\ \hline
\multicolumn{1}{l|}{\multirow{4}{*}{Ours}} & \multicolumn{2}{c|}{0.97} & \multicolumn{3}{c|}{1.40} & \multicolumn{1}{c|}{\multirow{4}{*}{0.04}} & \multirow{4}{*}{2.40} \\ \cline{2-6}
\multicolumn{1}{l|}{} & \multicolumn{1}{c|}{per-frame} & \multicolumn{1}{c|}{KF} & \multicolumn{1}{c|}{KF} & \multicolumn{1}{c|}{KF association+} & \multicolumn{1}{c|}{non-KF} & \multicolumn{1}{c|}{} &  \\
\multicolumn{1}{l|}{} & \multicolumn{1}{c|}{pre-process.} & \multicolumn{1}{c|}{decision} & \multicolumn{1}{c|}{optim.} & \multicolumn{1}{c|}{pose graph optim.} & \multicolumn{1}{c|}{optim.} & \multicolumn{1}{c|}{} &  \\
\multicolumn{1}{l|}{} & \multicolumn{1}{c|}{0.78} & \multicolumn{1}{c|}{0.19} & \multicolumn{1}{c|}{0.33} & \multicolumn{1}{c|}{0.01} & \multicolumn{1}{c|}{1.05} & \multicolumn{1}{c|}{} &  \\ \hline
\end{tabular}\vspace{0cm}\vspace{0.15cm}
    \begin{tabular}{lccccccr}
\multicolumn{8}{c}{\textbf{per-frame runtime on a 200-frame video}} \\ \hline
\multicolumn{1}{c|}{} & \multicolumn{2}{c|}{Pre-} & \multicolumn{3}{c|}{Main} & \multicolumn{1}{c|}{Post-} & \multicolumn{1}{c}{\multirow{2}{*}{Sum}} \\
\multicolumn{1}{c|}{} & \multicolumn{2}{c|}{process.} & \multicolumn{3}{c|}{procedure} & \multicolumn{1}{c|}{process.} & \multicolumn{1}{c}{} \\ \hline
\multicolumn{1}{l|}{Deep3D} & \multicolumn{2}{c|}{1.88} & \multicolumn{3}{c|}{1.14} & \multicolumn{1}{c|}{-} & 4.13 \\ \hline
\multicolumn{1}{l|}{CVD2} & \multicolumn{2}{c|}{1.61} & \multicolumn{3}{c|}{8.15} & \multicolumn{1}{c|}{5.07} & 14.83 \\ \hline
\multicolumn{1}{l|}{\multirow{4}{*}{Ours}} & \multicolumn{2}{c|}{0.97} & \multicolumn{3}{c|}{1.53} & \multicolumn{1}{c|}{\multirow{4}{*}{0.03}} & \multirow{4}{*}{2.53} \\ \cline{2-6}
\multicolumn{1}{l|}{} & \multicolumn{1}{c|}{per-frame} & \multicolumn{1}{c|}{KF} & \multicolumn{1}{c|}{KF} & \multicolumn{1}{c|}{KF association+} & \multicolumn{1}{c|}{non-KF} & \multicolumn{1}{c|}{} &  \\
\multicolumn{1}{l|}{} & \multicolumn{1}{c|}{pre-process.} & \multicolumn{1}{c|}{decision} & \multicolumn{1}{c|}{optim.} & \multicolumn{1}{c|}{pose graph optim.} & \multicolumn{1}{c|}{optim.} & \multicolumn{1}{c|}{} &  \\
\multicolumn{1}{l|}{} & \multicolumn{1}{c|}{0.67} & \multicolumn{1}{c|}{0.30} & \multicolumn{1}{c|}{0.29} & \multicolumn{1}{c|}{0.11} & \multicolumn{1}{c|}{1.12} & \multicolumn{1}{c|}{} &  \\ \hline
\end{tabular}\vspace{0cm}\vspace{0.15cm}
    \begin{tabular}{lccccccr}
\multicolumn{8}{c}{\textbf{per-frame runtime on a 500-frame video}} \\ \hline
\multicolumn{1}{c|}{} & \multicolumn{2}{c|}{Pre-} & \multicolumn{3}{c|}{Main} & \multicolumn{1}{c|}{Post-} & \multicolumn{1}{c}{\multirow{2}{*}{Sum}} \\
\multicolumn{1}{c|}{} & \multicolumn{2}{c|}{process.} & \multicolumn{3}{c|}{procedure} & \multicolumn{1}{c|}{process.} & \multicolumn{1}{c}{} \\ \hline
\multicolumn{1}{l|}{Deep3D} & \multicolumn{2}{c|}{1.90} & \multicolumn{3}{c|}{0.85} & \multicolumn{1}{c|}{-} & 2.75 \\ \hline
\multicolumn{1}{l|}{CVD2} & \multicolumn{2}{c|}{1.60} & \multicolumn{3}{c|}{10.69} & \multicolumn{1}{c|}{5.05} & 17.34 \\ \hline
\multicolumn{1}{l|}{\multirow{4}{*}{Ours}} & \multicolumn{2}{c|}{1.00} & \multicolumn{3}{c|}{1.71} & \multicolumn{1}{c|}{\multirow{4}{*}{0.03}} & \multirow{4}{*}{2.74} \\ \cline{2-6}
\multicolumn{1}{l|}{} & \multicolumn{1}{c|}{per-frame} & \multicolumn{1}{c|}{KF} & \multicolumn{1}{c|}{KF} & \multicolumn{1}{c|}{KF association+} & \multicolumn{1}{c|}{non-KF} & \multicolumn{1}{c|}{} &  \\
\multicolumn{1}{l|}{} & \multicolumn{1}{c|}{pre-process.} & \multicolumn{1}{c|}{decision} & \multicolumn{1}{c|}{optim.} & \multicolumn{1}{c|}{pose graph optim.} & \multicolumn{1}{c|}{optim.} & \multicolumn{1}{c|}{} &  \\
\multicolumn{1}{l|}{} & \multicolumn{1}{c|}{0.64} & \multicolumn{1}{c|}{0.36} & \multicolumn{1}{c|}{0.35} & \multicolumn{1}{c|}{0.11} & \multicolumn{1}{c|}{1.25} & \multicolumn{1}{c|}{} &  \\ \hline
\end{tabular}\vspace{0cm}\vspace{0.15cm}
    \begin{tabular}{lccccccr}
\multicolumn{8}{c}{\textbf{per-frame runtime on a 1000-frame video}} \\ \hline
\multicolumn{1}{c|}{} & \multicolumn{2}{c|}{Pre-} & \multicolumn{3}{c|}{Main} & \multicolumn{1}{c|}{Post-} & \multicolumn{1}{c}{\multirow{2}{*}{Sum}} \\
\multicolumn{1}{c|}{} & \multicolumn{2}{c|}{process.} & \multicolumn{3}{c|}{procedure} & \multicolumn{1}{c|}{process.} & \multicolumn{1}{c}{} \\ \hline
\multicolumn{1}{l|}{Deep3D} & \multicolumn{2}{c|}{1.91} & \multicolumn{3}{c|}{0.75} & \multicolumn{1}{c|}{-} & 2.66 \\ \hline
\multicolumn{1}{l|}{CVD2} & \multicolumn{2}{c|}{1.59} & \multicolumn{3}{c|}{15.07} & \multicolumn{1}{c|}{5.07} & 21.73 \\ \hline
\multicolumn{1}{l|}{\multirow{4}{*}{Ours}} & \multicolumn{2}{c|}{1.03} & \multicolumn{3}{c|}{1.77} & \multicolumn{1}{c|}{\multirow{4}{*}{0.03}} & \multirow{4}{*}{2.83} \\ \cline{2-6}
\multicolumn{1}{l|}{} & \multicolumn{1}{c|}{per-frame} & \multicolumn{1}{c|}{KF} & \multicolumn{1}{c|}{KF} & \multicolumn{1}{c|}{KF association+} & \multicolumn{1}{c|}{non-KF} & \multicolumn{1}{c|}{} &  \\
\multicolumn{1}{l|}{} & \multicolumn{1}{c|}{pre-process.} & \multicolumn{1}{c|}{decision} & \multicolumn{1}{c|}{optim.} & \multicolumn{1}{c|}{pose graph optim.} & \multicolumn{1}{c|}{optim.} & \multicolumn{1}{c|}{} &  \\
\multicolumn{1}{l|}{} & \multicolumn{1}{c|}{0.64} & \multicolumn{1}{c|}{0.39} & \multicolumn{1}{c|}{0.37} & \multicolumn{1}{c|}{0.13} & \multicolumn{1}{c|}{1.28} & \multicolumn{1}{c|}{} &  \\ \hline
\end{tabular}\vspace{0cm}\vspace{0.15cm}
    \begin{tabular}{lccccccr}
\multicolumn{8}{c}{\textbf{per-frame runtime on a 2000-frame video}} \\ \hline
\multicolumn{1}{c|}{} & \multicolumn{2}{c|}{Pre-} & \multicolumn{3}{c|}{Main} & \multicolumn{1}{c|}{Post-} & \multicolumn{1}{c}{\multirow{2}{*}{Sum}} \\
\multicolumn{1}{c|}{} & \multicolumn{2}{c|}{process.} & \multicolumn{3}{c|}{procedure} & \multicolumn{1}{c|}{process.} & \multicolumn{1}{c}{} \\ \hline
\multicolumn{1}{l|}{Deep3D} & \multicolumn{2}{c|}{1.90} & \multicolumn{3}{c|}{0.70} & \multicolumn{1}{c|}{-} & 2.61 \\ \hline
\multicolumn{1}{l|}{CVD2} & \multicolumn{2}{c|}{1.60} & \multicolumn{3}{c|}{19.44} & \multicolumn{1}{c|}{5.09} & 26.13 \\ \hline
\multicolumn{1}{l|}{\multirow{4}{*}{Ours}} & \multicolumn{2}{c|}{1.04} & \multicolumn{3}{c|}{1.96} & \multicolumn{1}{c|}{\multirow{4}{*}{0.03}} & \multirow{4}{*}{3.04} \\ \cline{2-6}
\multicolumn{1}{l|}{} & \multicolumn{1}{c|}{per-frame} & \multicolumn{1}{c|}{KF} & \multicolumn{1}{c|}{KF} & \multicolumn{1}{c|}{KF association+} & \multicolumn{1}{c|}{non-KF} & \multicolumn{1}{c|}{} &  \\
\multicolumn{1}{l|}{} & \multicolumn{1}{c|}{pre-process.} & \multicolumn{1}{c|}{decision} & \multicolumn{1}{c|}{optim.} & \multicolumn{1}{c|}{pose graph optim.} & \multicolumn{1}{c|}{optim.} & \multicolumn{1}{c|}{} &  \\
\multicolumn{1}{l|}{} & \multicolumn{1}{c|}{0.63} & \multicolumn{1}{c|}{0.41} & \multicolumn{1}{c|}{0.39} & \multicolumn{1}{c|}{0.25} & \multicolumn{1}{c|}{1.31} & \multicolumn{1}{c|}{} &  \\ \hline
\end{tabular}
    \caption{Runtime comparisons broken down by three main steps. The per-frame runtime are reported in seconds. (KF = keyframe)}
    \label{tab:time-step}
\end{table*}

\subsection{Runtime Analysis}\label{sec:runtime}
We analyze the runtime of each stage in our algorithm in detail. The pipeline is divided into three main steps, pre-processing, main procedure, and post-processing for fair comparison with Deep3D~\cite{lee20213d} and CVD2~\cite{kopf2021robust}.
The averaged per-frame runtimes of videos of varying length are presented in Table~\ref{tab:time-step}.  Note that Deep3D does not perform post-processing for video depth and pose estimation. 

In the pre-processing step, our method takes fewer time since we only requires adjacent optical flow. In contrast, Deep3D and CVD2 requires multiple pairs of optical flow for a target frame (\eg, $F_{t\pm\gamma\to t}, \gamma\in\{1,2,4,8\}$) an thus consumes more time.

In the main procedure for joint depth and pose optimization, although Deep3D takes fewer runtime by reducing the iterations for the optimization of non-first fragments, it may lead to sub-optimal results and yield global inconsistency. On the other hand, CVD2 consumes extremely long time due to the traditional optimization with CPU. The per-frame runtime of our main procedure slightly increase with the longer length of videos mainly due to the geometric verification in keyframe association and increasing non-sequential edges for pose graph optimization. Nevertheless, we emphasize the global consistency, especially in long videos. In the last post-processing step, we re-implement the post-processing proposed by CVD2 with GPU to shorten the computational time. 

In sum, the proposed method shows strong efficiency by using keyframe-based pose graph and optimization. Our globally-consistent method is slightly slower than Deep3D for the videos which is greater than 500 frames, yet is able to perform the fastest for the video less than 500 frames.

\subsection{Limitations}\label{sec:limitation}
Our method can achieve globally consistent depth and pose estimation with efficient test-time training. %, while 
Nevertheless, we discuss %the few 
the following cases that may introduce poor performance.

\noindent\textbf{Restriction by optical flow estimation.}
Although the dense optical flow estimation can improve the robustness of traditional sparse features, the performance of depth and pose substantially relies on the accurate optical flow. 
Yet, the state-of-the-art optical flow estimation could still suffer from the generalization issues and thus produce %poor 
un-satisfied flow estimation in some cases. 
Furthermore, the forward-backward consistency check is helpful but still cannot %essentially 
fully guarantee the accuracy of optical flow. 
Hence, %the further improvement of 
how to further improve the dense optical flow estimation remains a promising future direction. % is necessary.

\noindent\textbf{Learnable camera intrinsic parameters.}
%In our paper, 
In this work, we assume an ideal camera intrinsic with fixed focal length to simplify the learning of scale consistency in depth and camera pose. 
It is still challenging on handing the videos with varying focal lengths for global consistency.
Thus, we put the reconstruction of varying focal length in our future directions.

\clearpage

%\bibliographystyle{aaai22}
\bibliography{egbib}

\begin{thebibliography}{83}
\providecommand{\natexlab}[1]{#1}

\bibitem[{Agarwal, Mierle, and Others(2022)}]{ceres-solver}
Agarwal, S.; Mierle, K.; and Others. 2022.
\newblock Ceres Solver.
\newblock \url{http://ceres-solver.org}.

\bibitem[{Bao et~al.(2019)Bao, Lai, Ma, Zhang, Gao, and Yang}]{bao2019depth}
Bao, W.; Lai, W.-S.; Ma, C.; Zhang, X.; Gao, Z.; and Yang, M.-H. 2019.
\newblock Depth-aware video frame interpolation.
\newblock In \emph{CVPR}.

\bibitem[{Barath et~al.(2021)Barath, Mishkin, Eichhardt, Shipachev, and
  Matas}]{barath2021efficient}
Barath, D.; Mishkin, D.; Eichhardt, I.; Shipachev, I.; and Matas, J. 2021.
\newblock Efficient Initial Pose-graph Generation for Global SfM.
\newblock In \emph{CVPR}.

\bibitem[{Bay, Tuytelaars, and Van~Gool(2006)}]{bay2006surf}
Bay, H.; Tuytelaars, T.; and Van~Gool, L. 2006.
\newblock Surf: Speeded up robust features.
\newblock In \emph{ECCV}.

\bibitem[{Bian et~al.(2019)Bian, Li, Wang, Zhan, Shen, Cheng, and
  Reid}]{bian2019unsupervised}
Bian, J.; Li, Z.; Wang, N.; Zhan, H.; Shen, C.; Cheng, M.-M.; and Reid, I.
  2019.
\newblock Unsupervised scale-consistent depth and ego-motion learning from
  monocular video.
\newblock \emph{NeurIPS}.

\bibitem[{Bloesch et~al.(2018)Bloesch, Czarnowski, Clark, Leutenegger, and
  Davison}]{bloesch2018codeslam}
Bloesch, M.; Czarnowski, J.; Clark, R.; Leutenegger, S.; and Davison, A.~J.
  2018.
\newblock CodeSLAM—learning a compact, optimisable representation for dense
  visual SLAM.
\newblock In \emph{CVPR}.

\bibitem[{Burri et~al.(2016)Burri, Nikolic, Gohl, Schneider, Rehder, Omari,
  Achtelik, and Siegwart}]{Burri25012016}
Burri, M.; Nikolic, J.; Gohl, P.; Schneider, T.; Rehder, J.; Omari, S.;
  Achtelik, M.~W.; and Siegwart, R. 2016.
\newblock The EuRoC micro aerial vehicle datasets.
\newblock \emph{The International Journal of Robotics Research (IJRR)}.

\bibitem[{Casser et~al.(2019)Casser, Pirk, Mahjourian, and
  Angelova}]{casser2019unsupervised}
Casser, V.; Pirk, S.; Mahjourian, R.; and Angelova, A. 2019.
\newblock Depth prediction without the sensors: Leveraging structure for
  unsupervised learning from monocular videos.
\newblock In \emph{AAAI}.

\bibitem[{Chen et~al.(2016)Chen, Fu, Yang, and Deng}]{chen2016single}
Chen, W.; Fu, Z.; Yang, D.; and Deng, J. 2016.
\newblock Single-image depth perception in the wild.
\newblock \emph{NeurIPS}.

\bibitem[{Chen et~al.(2020)Chen, Qian, Fan, Kojima, Hamilton, and
  Deng}]{chen2020oasis}
Chen, W.; Qian, S.; Fan, D.; Kojima, N.; Hamilton, M.; and Deng, J. 2020.
\newblock Oasis: A large-scale dataset for single image 3d in the wild.
\newblock In \emph{CVPR}.

\bibitem[{Chen, Schmid, and Sminchisescu(2019)}]{chen2019self}
Chen, Y.; Schmid, C.; and Sminchisescu, C. 2019.
\newblock Self-supervised learning with geometric constraints in monocular
  video: Connecting flow, depth, and camera.
\newblock In \emph{ICCV}.

\bibitem[{Choi et~al.(2019)Choi, Gallo, Troccoli, Kim, and
  Kautz}]{choi2019extreme}
Choi, I.; Gallo, O.; Troccoli, A.; Kim, M.~H.; and Kautz, J. 2019.
\newblock Extreme view synthesis.
\newblock In \emph{CVPR}.

\bibitem[{Czarnowski et~al.(2020)Czarnowski, Laidlow, Clark, and
  Davison}]{czarnowski2020deepfactors}
Czarnowski, J.; Laidlow, T.; Clark, R.; and Davison, A.~J. 2020.
\newblock Deepfactors: Real-time probabilistic dense monocular slam.
\newblock \emph{IEEE Robotics and Automation Letters}.

\bibitem[{Du et~al.(2020)Du, Turner, Dzitsiuk, Prasso, Duarte, Dourgarian,
  Afonso, Pascoal, Gladstone, Cruces, Izadi, Kowdle, Tsotsos, and
  Kim}]{Du2020DepthLab}
Du, R.; Turner, E.; Dzitsiuk, M.; Prasso, L.; Duarte, I.; Dourgarian, J.;
  Afonso, J.; Pascoal, J.; Gladstone, J.; Cruces, N.; Izadi, S.; Kowdle, A.;
  Tsotsos, K.; and Kim, D. 2020.
\newblock {DepthLab: Real-Time 3D Interaction With Depth Maps for Mobile
  Augmented Reality}.
\newblock In \emph{Proceedings of the 33rd Annual ACM Symposium on User
  Interface Software and Technology}, UIST. ACM.

\bibitem[{Eigen and Fergus(2015)}]{eigen2015predicting}
Eigen, D.; and Fergus, R. 2015.
\newblock Predicting depth, surface normals and semantic labels with a common
  multi-scale convolutional architecture.
\newblock In \emph{ICCV}.

\bibitem[{Eigen, Puhrsch, and Fergus(2014)}]{eigen2014depth}
Eigen, D.; Puhrsch, C.; and Fergus, R. 2014.
\newblock Depth Map Prediction from a Single Image using a Multi-Scale Deep
  Network.
\newblock In \emph{NeurIPS}. Curran Associates, Inc.

\bibitem[{Engel, Sch{\"o}ps, and Cremers(2014)}]{engel2014lsd}
Engel, J.; Sch{\"o}ps, T.; and Cremers, D. 2014.
\newblock LSD-SLAM: Large-scale direct monocular SLAM.
\newblock In \emph{ECCV}.

\bibitem[{et~al.(2017)}]{tateno2017cnn}
et~al., K.~T. 2017.
\newblock Cnn-slam: Real-time dense monocular slam with learned depth
  prediction.
\newblock In \emph{CVPR}.

\bibitem[{Fischler and Bolles(1981)}]{fischler1981random}
Fischler, M.~A.; and Bolles, R.~C. 1981.
\newblock Random sample consensus: a paradigm for model fitting with
  applications to image analysis and automated cartography.
\newblock \emph{Communications of the ACM}.

\bibitem[{Furukawa and Hern\'{a}ndez(2015)}]{furukawa2015mvstutorial}
Furukawa, Y.; and Hern\'{a}ndez, C. 2015.
\newblock Multi-View Stereo: A Tutorial.
\newblock \emph{Found. Trends. Comput. Graph. Vis.}

\bibitem[{Garg et~al.(2016)Garg, Bg, Carneiro, and Reid}]{garg2016unsupervised}
Garg, R.; Bg, V.~K.; Carneiro, G.; and Reid, I. 2016.
\newblock Unsupervised cnn for single view depth estimation: Geometry to the
  rescue.
\newblock In \emph{ECCV}.

\bibitem[{Geiger et~al.(2013)Geiger, Lenz, Stiller, and
  Urtasun}]{Geiger2013IJRR}
Geiger, A.; Lenz, P.; Stiller, C.; and Urtasun, R. 2013.
\newblock Vision meets Robotics: The KITTI Dataset.
\newblock \emph{International Journal of Robotics Research (IJRR)}.

\bibitem[{Godard, Mac~Aodha, and Brostow(2017)}]{godard2017unsupervised}
Godard, C.; Mac~Aodha, O.; and Brostow, G.~J. 2017.
\newblock Unsupervised monocular depth estimation with left-right consistency.
\newblock In \emph{CVPR}.

\bibitem[{Godard et~al.(2019)Godard, Mac~Aodha, Firman, and
  Brostow}]{godard2019digging}
Godard, C.; Mac~Aodha, O.; Firman, M.; and Brostow, G.~J. 2019.
\newblock Digging into self-supervised monocular depth estimation.
\newblock In \emph{ICCV}.

\bibitem[{Gonzalez and Kim(2021)}]{gonzalez2021plade}
Gonzalez, J.~L.; and Kim, M. 2021.
\newblock PLADE-Net: Towards Pixel-Level Accuracy for Self-Supervised
  Single-View Depth Estimation With Neural Positional Encoding and Distilled
  Matting Loss.
\newblock In \emph{CVPR}.

\bibitem[{Gordon et~al.(2019)Gordon, Li, Jonschkowski, and
  Angelova}]{gordon2019depth}
Gordon, A.; Li, H.; Jonschkowski, R.; and Angelova, A. 2019.
\newblock Depth from videos in the wild: Unsupervised monocular depth learning
  from unknown cameras.
\newblock In \emph{ICCV}.

\bibitem[{Hartley and Zisserman(2003)}]{hartley2003multiple}
Hartley, R.; and Zisserman, A. 2003.
\newblock \emph{Multiple view geometry in computer vision}.
\newblock Cambridge university press.

\bibitem[{He et~al.(2016)He, Zhang, Ren, and Sun}]{he2016deep}
He, K.; Zhang, X.; Ren, S.; and Sun, J. 2016.
\newblock Deep residual learning for image recognition.
\newblock In \emph{CVPR}.

\bibitem[{Holynski and Kopf(2018)}]{Occlusion2018}
Holynski, A.; and Kopf, J. 2018.
\newblock Fast Depth Densification for Occlusion-aware Augmented Reality.
\newblock 37(6).

\bibitem[{Huang et~al.(2018)Huang, Matzen, Kopf, Ahuja, and
  Huang}]{huang2018deepmvs}
Huang, P.-H.; Matzen, K.; Kopf, J.; Ahuja, N.; and Huang, J.-B. 2018.
\newblock Deepmvs: Learning multi-view stereopsis.
\newblock In \emph{CVPR}.

\bibitem[{Im et~al.(2019)Im, Jeon, Lin, and Kweon}]{im2019dpsnet}
Im, S.; Jeon, H.-G.; Lin, S.; and Kweon, I.~S. 2019.
\newblock Dpsnet: End-to-end deep plane sweep stereo.
\newblock \emph{arXiv preprint arXiv:1905.00538}.

\bibitem[{Jiao, Tran, and Shi(2021)}]{jiao2021effiscene}
Jiao, Y.; Tran, T.~D.; and Shi, G. 2021.
\newblock EffiScene: Efficient Per-Pixel Rigidity Inference for Unsupervised
  Joint Learning of Optical Flow, Depth, Camera Pose and Motion Segmentation.
\newblock In \emph{CVPR}.

\bibitem[{Jung, Park, and Yoo(2021)}]{Jung_2021_ICCV}
Jung, H.; Park, E.; and Yoo, S. 2021.
\newblock Fine-Grained Semantics-Aware Representation Enhancement for
  Self-Supervised Monocular Depth Estimation.
\newblock In \emph{ICCV}.

\bibitem[{Klein and Murray(2007)}]{klein2007parallel}
Klein, G.; and Murray, D. 2007.
\newblock Parallel tracking and mapping for small AR workspaces.
\newblock In \emph{2007 6th IEEE and ACM International Symposium on Mixed and
  Augmented Reality}.

\bibitem[{Kopf, Rong, and Huang(2021)}]{kopf2021robust}
Kopf, J.; Rong, X.; and Huang, J.-B. 2021.
\newblock Robust consistent video depth estimation.
\newblock In \emph{CVPR}.

\bibitem[{K{\"u}mmerle et~al.(2011)K{\"u}mmerle, Grisetti, Strasdat, Konolige,
  and Burgard}]{kummerle2011g}
K{\"u}mmerle, R.; Grisetti, G.; Strasdat, H.; Konolige, K.; and Burgard, W.
  2011.
\newblock g2o: A general framework for graph optimization.
\newblock In \emph{ICRA}.

\bibitem[{Lee et~al.(2021)Lee, Tseng, Chen, Chen, Chen, and Hung}]{lee20213d}
Lee, Y.-C.; Tseng, K.-W.; Chen, Y.-T.; Chen, C.-C.; Chen, C.-S.; and Hung,
  Y.-P. 2021.
\newblock 3D Video Stabilization With Depth Estimation by CNN-Based
  Optimization.
\newblock In \emph{CVPR}.

\bibitem[{Li et~al.(2019{\natexlab{a}})Li, Xue, Wang, Yan, and
  Zha}]{li2019sequential}
Li, S.; Xue, F.; Wang, X.; Yan, Z.; and Zha, H. 2019{\natexlab{a}}.
\newblock Sequential adversarial learning for self-supervised deep visual
  odometry.
\newblock In \emph{ICCV}.

\bibitem[{Li et~al.(2019{\natexlab{b}})Li, Dekel, Cole, Tucker, Snavely, Liu,
  and Freeman}]{li2019learning}
Li, Z.; Dekel, T.; Cole, F.; Tucker, R.; Snavely, N.; Liu, C.; and Freeman,
  W.~T. 2019{\natexlab{b}}.
\newblock Learning the depths of moving people by watching frozen people.
\newblock In \emph{CVPR}.

\bibitem[{Li and Snavely(2018)}]{li2018megadepth}
Li, Z.; and Snavely, N. 2018.
\newblock Megadepth: Learning single-view depth prediction from internet
  photos.
\newblock In \emph{CVPR}.

\bibitem[{Liu et~al.(2021)Liu, Tucker, Jampani, Makadia, Snavely, and
  Kanazawa}]{liu2021infinite}
Liu, A.; Tucker, R.; Jampani, V.; Makadia, A.; Snavely, N.; and Kanazawa, A.
  2021.
\newblock Infinite nature: Perpetual view generation of natural scenes from a
  single image.
\newblock In \emph{ICCV}.

\bibitem[{Liu et~al.(2009)Liu, Gleicher, Jin, and Agarwala}]{liu2009content}
Liu, F.; Gleicher, M.; Jin, H.; and Agarwala, A. 2009.
\newblock Content-preserving warps for 3D video stabilization.
\newblock \emph{ACM TOG}.

\bibitem[{Liu et~al.(2015)Liu, Shen, Lin, and Reid}]{liu2015learning}
Liu, F.; Shen, C.; Lin, G.; and Reid, I. 2015.
\newblock Learning depth from single monocular images using deep convolutional
  neural fields.
\newblock \emph{IEEE TPAMI}.

\bibitem[{Long et~al.(2021)Long, Liu, Li, Theobalt, and Wang}]{long2021multi}
Long, X.; Liu, L.; Li, W.; Theobalt, C.; and Wang, W. 2021.
\newblock Multi-view Depth Estimation using Epipolar Spatio-Temporal Networks.
\newblock In \emph{CVPR}.

\bibitem[{Long et~al.(2020)Long, Liu, Theobalt, and Wang}]{long2020occlusion}
Long, X.; Liu, L.; Theobalt, C.; and Wang, W. 2020.
\newblock Occlusion-aware depth estimation with adaptive normal constraints.
\newblock In \emph{ECCV}.

\bibitem[{Lowe(2004)}]{lowe2004distinctive}
Lowe, D.~G. 2004.
\newblock Distinctive image features from scale-invariant keypoints.
\newblock \emph{IJCV}.

\bibitem[{Luo et~al.(2020)Luo, Huang, Szeliski, Matzen, and
  Kopf}]{luo2020consistent}
Luo, X.; Huang, J.-B.; Szeliski, R.; Matzen, K.; and Kopf, J. 2020.
\newblock Consistent video depth estimation.
\newblock \emph{ACM TOG}.

\bibitem[{Mayer et~al.(2016)Mayer, Ilg, Hausser, Fischer, Cremers, Dosovitskiy,
  and Brox}]{mayer2016large}
Mayer, N.; Ilg, E.; Hausser, P.; Fischer, P.; Cremers, D.; Dosovitskiy, A.; and
  Brox, T. 2016.
\newblock A large dataset to train convolutional networks for disparity,
  optical flow, and scene flow estimation.
\newblock In \emph{CVPR}.

\bibitem[{Miangoleh et~al.(2021)Miangoleh, Dille, Mai, Paris, and
  Aksoy}]{miangoleh2021boosting}
Miangoleh, S. M.~H.; Dille, S.; Mai, L.; Paris, S.; and Aksoy, Y. 2021.
\newblock Boosting Monocular Depth Estimation Models to High-Resolution via
  Content-Adaptive Multi-Resolution Merging.
\newblock In \emph{CVPR}.

\bibitem[{Moulon et~al.(2016)Moulon, Monasse, Perrot, and
  Marlet}]{moulon2016openmvg}
Moulon, P.; Monasse, P.; Perrot, R.; and Marlet, R. 2016.
\newblock Openmvg: Open multiple view geometry.
\newblock In \emph{International Workshop on Reproducible Research in Pattern
  Recognition}.

\bibitem[{Mur-Artal, Montiel, and Tardos(2015)}]{mur2015orb}
Mur-Artal, R.; Montiel, J. M.~M.; and Tardos, J.~D. 2015.
\newblock ORB-SLAM: a versatile and accurate monocular SLAM system.
\newblock \emph{IEEE Transactions on Robotics}.

\bibitem[{Nathan~Silberman and Fergus(2012)}]{Silberman:ECCV12}
Nathan~Silberman, P.~K., Derek~Hoiem; and Fergus, R. 2012.
\newblock Indoor Segmentation and Support Inference from RGBD Images.
\newblock In \emph{ECCV}.

\bibitem[{Ranftl et~al.(2020)Ranftl, Lasinger, Hafner, Schindler, and
  Koltun}]{Ranftl2020}
Ranftl, R.; Lasinger, K.; Hafner, D.; Schindler, K.; and Koltun, V. 2020.
\newblock Towards Robust Monocular Depth Estimation: Mixing Datasets for
  Zero-shot Cross-dataset Transfer.
\newblock \emph{IEEE TPAMI}.

\bibitem[{Ranjan et~al.(2019)Ranjan, Jampani, Balles, Kim, Sun, Wulff, and
  Black}]{ranjan2019competitive}
Ranjan, A.; Jampani, V.; Balles, L.; Kim, K.; Sun, D.; Wulff, J.; and Black,
  M.~J. 2019.
\newblock Competitive collaboration: Joint unsupervised learning of depth,
  camera motion, optical flow and motion segmentation.
\newblock In \emph{CVPR}.

\bibitem[{Rublee et~al.(2011)Rublee, Rabaud, Konolige, and
  Bradski}]{rublee2011orb}
Rublee, E.; Rabaud, V.; Konolige, K.; and Bradski, G. 2011.
\newblock ORB: An efficient alternative to SIFT or SURF.
\newblock In \emph{ICCV}.

\bibitem[{Saxena, Sun, and Ng(2008)}]{saxena2008make3d}
Saxena, A.; Sun, M.; and Ng, A.~Y. 2008.
\newblock Make3d: Learning 3d scene structure from a single still image.
\newblock \emph{IEEE TPAMI}.

\bibitem[{Schonberger and Frahm(2016)}]{schonberger2016structure}
Schonberger, J.~L.; and Frahm, J.-M. 2016.
\newblock Structure-from-motion revisited.
\newblock In \emph{CVPR}.

\bibitem[{Shotton et~al.(2013)Shotton, Glocker, Zach, Izadi, Criminisi, and
  Fitzgibbon}]{shotton2013scene}
Shotton, J.; Glocker, B.; Zach, C.; Izadi, S.; Criminisi, A.; and Fitzgibbon,
  A. 2013.
\newblock Scene coordinate regression forests for camera relocalization in
  RGB-D images.
\newblock In \emph{CVPR}.

\bibitem[{Sturm et~al.(2012)Sturm, Engelhard, Endres, Burgard, and
  Cremers}]{sturm12iros}
Sturm, J.; Engelhard, N.; Endres, F.; Burgard, W.; and Cremers, D. 2012.
\newblock A Benchmark for the Evaluation of RGB-D SLAM Systems.
\newblock In \emph{IEEE/RSJ Int. Conf. Intell. Robot. Syst. (IROS)}.

\bibitem[{Sun et~al.(2021)Sun, Xie, Chen, Zhou, and Bao}]{sun2021neuralrecon}
Sun, J.; Xie, Y.; Chen, L.; Zhou, X.; and Bao, H. 2021.
\newblock NeuralRecon: Real-Time Coherent 3D Reconstruction from Monocular
  Video.
\newblock In \emph{CVPR}.

\bibitem[{Teed and Deng(2019)}]{teed2018deepv2d}
Teed, Z.; and Deng, J. 2019.
\newblock DeepV2D: Video to Depth with Differentiable Structure from Motion.
\newblock In \emph{International Conference on Learning Representations}.

\bibitem[{Teed and Deng(2020)}]{teed2020raft}
Teed, Z.; and Deng, J. 2020.
\newblock Raft: Recurrent all-pairs field transforms for optical flow.
\newblock In \emph{ECCV}.

\bibitem[{Teed and Deng(2021)}]{teed2021droid}
Teed, Z.; and Deng, J. 2021.
\newblock Droid-slam: Deep visual slam for monocular, stereo, and rgb-d
  cameras.
\newblock \emph{NeurIPS}.

\bibitem[{Triggs et~al.(1999)Triggs, McLauchlan, Hartley, and
  Fitzgibbon}]{triggs1999bundle}
Triggs, B.; McLauchlan, P.~F.; Hartley, R.~I.; and Fitzgibbon, A.~W. 1999.
\newblock Bundle adjustment—a modern synthesis.
\newblock In \emph{International workshop on vision algorithms}. Springer.

\bibitem[{Ummenhofer et~al.(2017)Ummenhofer, Zhou, Uhrig, Mayer, Ilg,
  Dosovitskiy, and Brox}]{ummenhofer2017demon}
Ummenhofer, B.; Zhou, H.; Uhrig, J.; Mayer, N.; Ilg, E.; Dosovitskiy, A.; and
  Brox, T. 2017.
\newblock Demon: Depth and motion network for learning monocular stereo.
\newblock In \emph{CVPR}.

\bibitem[{Vaswani et~al.(2017)Vaswani, Shazeer, Parmar, Uszkoreit, Jones,
  Gomez, Kaiser, and Polosukhin}]{vaswani2017attention}
Vaswani, A.; Shazeer, N.; Parmar, N.; Uszkoreit, J.; Jones, L.; Gomez, A.~N.;
  Kaiser, {\L}.; and Polosukhin, I. 2017.
\newblock Attention is all you need.
\newblock In \emph{NeurIPS}.

\bibitem[{Wang et~al.(2021)Wang, Zhong, Dai, Birchfield, Zhang, Smolyanskiy,
  and Li}]{wang2021deep}
Wang, J.; Zhong, Y.; Dai, Y.; Birchfield, S.; Zhang, K.; Smolyanskiy, N.; and
  Li, H. 2021.
\newblock Deep Two-View Structure-from-Motion Revisited.
\newblock In \emph{CVPR}.

\bibitem[{Wang et~al.(2004)Wang, Bovik, Sheikh, and Simoncelli}]{wang2004image}
Wang, Z.; Bovik, A.~C.; Sheikh, H.~R.; and Simoncelli, E.~P. 2004.
\newblock Image quality assessment: from error visibility to structural
  similarity.
\newblock \emph{IEEE TIP}.

\bibitem[{Watson et~al.(2021)Watson, Mac~Aodha, Prisacariu, Brostow, and
  Firman}]{watson2021temporal}
Watson, J.; Mac~Aodha, O.; Prisacariu, V.; Brostow, G.; and Firman, M. 2021.
\newblock The Temporal Opportunist: Self-Supervised Multi-Frame Monocular
  Depth.
\newblock In \emph{CVPR}.

\bibitem[{Wei et~al.(2020)Wei, Zhang, Li, Fu, and Xue}]{wei2020deepsfm}
Wei, X.; Zhang, Y.; Li, Z.; Fu, Y.; and Xue, X. 2020.
\newblock Deepsfm: Structure from motion via deep bundle adjustment.
\newblock In \emph{ECCV}.

\bibitem[{Wimbauer et~al.(2021)Wimbauer, Yang, von Stumberg, Zeller, and
  Cremers}]{wimbauer2021monorec}
Wimbauer, F.; Yang, N.; von Stumberg, L.; Zeller, N.; and Cremers, D. 2021.
\newblock MonoRec: Semi-supervised dense reconstruction in dynamic environments
  from a single moving camera.
\newblock In \emph{CVPR}.

\bibitem[{Wu et~al.(2011)}]{wu2011visualsfm}
Wu, C.; et~al. 2011.
\newblock VisualSFM: A visual structure from motion system.

\bibitem[{Wu et~al.(2019)Wu, Kirillov, Massa, Lo, and
  Girshick}]{wu2019detectron2}
Wu, Y.; Kirillov, A.; Massa, F.; Lo, W.-Y.; and Girshick, R. 2019.
\newblock Detectron2.
\newblock \url{https://github.com/facebookresearch/detectron2}.

\bibitem[{Wulff et~al.(2012)Wulff, Butler, Stanley, and
  Black}]{Wulff:ECCVws:2012}
Wulff, J.; Butler, D.~J.; Stanley, G.~B.; and Black, M.~J. 2012.
\newblock Lessons and insights from creating a synthetic optical flow
  benchmark.
\newblock In \emph{ECCV Workshop on Unsolved Problems in Optical Flow and
  Stereo Estimation}.

\bibitem[{Yao et~al.(2018)Yao, Luo, Li, Fang, and Quan}]{yao2018mvsnet}
Yao, Y.; Luo, Z.; Li, S.; Fang, T.; and Quan, L. 2018.
\newblock Mvsnet: Depth inference for unstructured multi-view stereo.
\newblock In \emph{ECCV}.

\bibitem[{Yin et~al.(2021)Yin, Zhang, Wang, Niklaus, Mai, Chen, and
  Shen}]{yin2021learning}
Yin, W.; Zhang, J.; Wang, O.; Niklaus, S.; Mai, L.; Chen, S.; and Shen, C.
  2021.
\newblock Learning to recover 3d scene shape from a single image.
\newblock In \emph{CVPR}.

\bibitem[{Yin and Shi(2018)}]{yin2018geonet}
Yin, Z.; and Shi, J. 2018.
\newblock Geonet: Unsupervised learning of dense depth, optical flow and camera
  pose.
\newblock In \emph{CVPR}.

\bibitem[{Yoon et~al.(2020)Yoon, Kim, Gallo, Park, and Kautz}]{yoon2020novel}
Yoon, J.~S.; Kim, K.; Gallo, O.; Park, H.~S.; and Kautz, J. 2020.
\newblock Novel view synthesis of dynamic scenes with globally coherent depths
  from a monocular camera.
\newblock In \emph{CVPR}.

\bibitem[{Zhang and Scaramuzza(2018)}]{Zhang18iros}
Zhang, Z.; and Scaramuzza, D. 2018.
\newblock A Tutorial on Quantitative Trajectory Evaluation for
  Visual(-Inertial) Odometry.
\newblock In \emph{IEEE/RSJ Int. Conf. Intell. Robot. Syst. (IROS)}.

\bibitem[{Zhou, Ummenhofer, and Brox(2018)}]{zhou2018deeptam}
Zhou, H.; Ummenhofer, B.; and Brox, T. 2018.
\newblock Deeptam: Deep tracking and mapping.
\newblock In \emph{ECCV}.

\bibitem[{Zhou et~al.(2017)Zhou, Brown, Snavely, and
  Lowe}]{zhou2017unsupervised}
Zhou, T.; Brown, M.; Snavely, N.; and Lowe, D.~G. 2017.
\newblock Unsupervised learning of depth and ego-motion from video.
\newblock In \emph{CVPR}.

\bibitem[{Zhou et~al.(2021)Zhou, Fan, Shi, and Xin}]{Zhou_2021_ICCV}
Zhou, Z.; Fan, X.; Shi, P.; and Xin, Y. 2021.
\newblock R-MSFM: Recurrent Multi-Scale Feature Modulation for Monocular Depth
  Estimating.
\newblock In \emph{ICCV}.

\bibitem[{Zou et~al.(2020)Zou, Ji, Tran, Huang, and
  Chandraker}]{zou2020learning}
Zou, Y.; Ji, P.; Tran, Q.-H.; Huang, J.-B.; and Chandraker, M. 2020.
\newblock Learning monocular visual odometry via self-supervised long-term
  modeling.
\newblock In \emph{ECCV}.

\end{thebibliography}
\end{document}